\documentclass{article}
\usepackage[preprint]{neurips_2023}

\usepackage[utf8]{inputenc} 
\usepackage[T1]{fontenc}    
\usepackage{hyperref}       
\usepackage{url}            
\usepackage{booktabs}       
\usepackage{amsfonts}       
\usepackage{nicefrac}       
\usepackage{microtype}      
\usepackage{xcolor}         
\usepackage{graphicx}
\usepackage{subcaption}
\title{Effective Latent Differential Equation Models \\ via Attention and Multiple Shooting}

\usepackage{authblk}

\usepackage{anyfontsize}

\newsavebox\affbox

\author[1, *]{Germán Abrevaya}
\author[2, 3]{Mahta Ramezanian-Panahi}
\author[2]{Jean-Christophe Gagnon-Audet}
\author[2, 3]{\newline Irina Rish}
\author[4]{Pablo Polosecki}
\author[1, 5]{Silvina Ponce Dawson}
\author[4]{Guillermo Cecchi}
\author[2, 6, 7]{\newline Guillaume Dumas}

\affil[1]{Universidad de Buenos Aires, Facultad de Ciencias Exactas y Naturales, Departamento de Física. Buenos Aires, Argentina.}
\affil[2]{Mila - Quebec AI Institute. Montréal, Québec, Canada.}
\affil[3]{University of Montreal. Montréal, Québec, Canada.}
\affil[4]{Healthcare and Life Sciences, IBM Research, T.J. Watson Research Center. Yorktown Heights, New York, USA.}
\affil[5]{CONICET - Universidad de Buenos Aires, Instituto de Física de Buenos Aires (IFIBA). Buenos Aires, Argentina.}
\affil[6]{Research Center of the Sainte-Justine Mother and Child University Hospital Center (CHU Sainte-Justine). Montréal, Québec, Canada.}
\affil[7]{Department of Psychiatry and Addictology, University of Montreal. Montréal, Québec, Canada.}
\affil[*]{\href{mailto:gabrevaya@df.uba.ar}{Correspondence to \texttt{gabrevaya@df.uba.ar}}}

\begin{document}

\maketitle

\begin{abstract}
Scientific Machine Learning (SciML) is a burgeoning field that synergistically combines domain-aware and interpretable models with agnostic machine learning techniques. In this work, we introduce GOKU-UI, an evolution of the SciML generative model GOKU-nets. GOKU-UI not only broadens the original model's spectrum to incorporate other classes of differential equations, such as Stochastic Differential Equations (SDEs), but also integrates attention mechanisms and a novel multiple shooting training strategy in the latent space. These modifications have led to a significant increase in its performance in both reconstruction and forecast tasks, as demonstrated by our evaluation of simulated and empirical data. Specifically, GOKU-UI outperformed all baseline models on synthetic datasets even with a training set 16-fold smaller, underscoring its remarkable data efficiency. Furthermore, when applied to empirical human brain data, while incorporating stochastic Stuart-Landau oscillators into its dynamical core, our proposed enhancements markedly increased the model's effectiveness in capturing complex brain dynamics. This augmented version not only surpassed all baseline methods in the reconstruction task, but also demonstrated lower prediction error of future brain activity up to 15 seconds ahead. By training GOKU-UI on resting state fMRI data, we encoded whole-brain dynamics into a latent representation, learning a low-dimensional dynamical system model that could offer insights into brain functionality and open avenues for practical applications such as the classification of mental states or psychiatric conditions. Ultimately, our research provides further impetus for the field of Scientific Machine Learning, showcasing the potential for advancements when established scientific insights are interwoven with modern machine learning.
\end{abstract}

\section{Introduction}
\label{intro}
\subsection{Scientific Machine Learning}
Scientific Machine Learning (SciML) is an emerging field that, drawing insights from scientific data, seeks to advance data-driven discovery with an approach that can produce interpretable results~\citep{osti_1478744}. Its synergistic blend of machine learning and scientific computing based on mechanistic models makes it very powerful for addressing complex problems across all STEM areas and beyond~\citep{willard_2022}. Using informed priors, its application is already making key contributions in scientific inference, data analysis, and machine learning enhanced modeling~\citep{von_rueden_2021}. Recent developments in SciML include various approaches to derive dynamical system models from observational data. The sparse identification of nonlinear dynamics (SINDy) algorithm is one such approach which, leveraging recent advances in sparsity techniques, exploits the observation that only a few important terms dominate the dynamics in most physical systems~\citep{brunton_2016}. A different well-established candidate for handling model-agnostic systems is deep reinforcement learning, which has been particularly widely adopted by the fluid mechanics community, with applications in flow control and shape optimization~\citep{martin2021reinforcement, viquerat2022review}. Physics-informed neural networks (PINNs) are another approach in which neural networks are trained to solve supervised learning tasks with respect to the laws of physics and can be used to derive data-driven partial differential equations or their solutions~\citep{RAISSI2019686}.  The application of PINNs to numerically stiff systems~\citep{wang_siam_2021} is challenging. Universal differential equations (UDEs) are a recent method that not only overcomes the stiffness limitation, but also represents a perfect example of the essence of SciML: use all the prior knowledge and scientific insights available for your problem and fill the missing gaps with machine learning~\citep{rackauckas_2020}. The simple yet powerful idea behind UDEs involves using traditional differential equation models with some unknown terms that are substituted by universal function approximators, such as neural networks. These approximators will be learned simultaneously with the equation’s parameters by using sensitivity algorithms~\citep{ma2021comparison} within the differentiable modelling framework~\citep{shen2023differentiable}.
  
The evolution equations that might be derived from the data may not actually correspond to mechanistic models based on first principles. That is, on many occasions, the dynamics take place on a manifold of a lower dimension than the full phase space of the system~\citep{FOIAS1988309}. Having reduced equations that describe the evolution on these manifolds is very useful, especially in high-dimensional systems~\citep{LUCIA200451}. Reduced-order models (ROMs) can be derived from data~\citep{GUO201975}. In particular, generative adversarial network (GAN)~\citep{GAN_NIPS2014} approaches have been used to enhance the application of ROMs to simulations of fluid dynamics~\citep{kim_2019,kim_2021}.
  
\subsection{Neural Differential Equations}
Despite the basic ideas behind differential equations parameterized by neural networks and its connection with deep learning had older roots in literature~\citep{rico1992discrete, rico1994continuous, rico1993continuous, chang2018reversible, weinan2017proposal, haber2017stable, lu2018beyond}, the publication of~\citet{chen_2018} was a turning point in the young history of SciML. Since then, the topic of \emph{neural differential equations} (neural DEs) has become a field, as stated and evidenced in the comprehensive survey by~\citet{kidger2022neural}. In~\citet{chen_2018}, by interpreting ResNets~\citep{he2016deep} as a discrete integration of a vector field with the Euler method, the authors proposed an infinitesimally layered neural network as its continuous limit and modeled it with an ordinary differential equation (ODE) parameterized by a neural network, giving rise to the Neural Ordinary Differential Equation (NODE) models. They also demonstrated that NODEs can be trained by backpropagating through black-box ODE solvers using the adjoint method~\citep{pearlmutter1995gradient}, making it a memory-efficient model.

Furthermore,~\citet{chen_2018} introduced the Latent Ordinary Differential Equations (Latent ODEs), a continuous-time generative model that encodes time series data into a latent space that could potentially capture its underlying dynamics, which are modeled using a NODE. First, the observed time series are encoded in the latent space using a recognition model, typically a Recurrent Neural Network (RNN). The temporal dynamics in the latent space is then modeled using NODEs, and lastly, its solution is decoded back into the observation space to generate predictions or perform other tasks such as anomaly detection or imputation of missing values.

By using this approach, Latent ODEs can capture the intricate and potentially nonlinear dynamical systems that underlie many real-world time series data. The continuous nature of NODEs allows the model to handle irregularly sampled data~\citep{rubanova2019latent}, a common challenge in many applications. Additionally, the use of the latent space enables the model to capture complex patterns in high-dimensional data, while still providing a compact and interpretable representation of the underlying dynamics.

\subsection{GOKU-nets}
The work of~\citet{Linial_2021} builds on the basis of the Latent ODEs model demonstrating that the incorporation of prior knowledge of the dynamics involved in the form of a backbone differential equation structure can increase the performance of a purely agnostic model. They propose another continuous-time generative model called GOKU-nets (which stands for Generative ODE Modeling with Known Unknowns), which are the focus of this paper. This model incorporates a variational autoencoder (VAE) structure with a differential equation to model the dynamics in the latent space. However, in this case, a specific form for the ODE is provided while allowing its parameters (the \emph{known unknowns}) to be inferred. The model is trained end-to-end jointly learning the transformation to the latent space, inferring the initial conditions and parameters of the ODE, which will be integrated afterward; and finally, a last transformation is performed to go back to the input space. In the next section, the details of its architecture will be described in detail. Note that the ODE can be integrated further than the input time span in order to generate an extrapolation, thus becoming a forecast of the future evolution of the time series. The study in~\citet{Linial_2021} compares GOKU-net with baselines such as LSTM and Latent-ODE in three domains: a video of a pendulum, a video of a double pendulum, and a dynamic model of the cardiovascular system. The authors show that their model significantly outperforms the others in reconstruction and extrapolation capabilities, reduces the size of the required training sets for effective learning, and furthermore, has greater interpretability, allowing for the identification of unobserved but clinically meaningful parameters.

The original GOKU-net model was limited to handling only ODEs. In this work, we expand its capabilities by implementing the model in the Julia Programming Language~\citep{bezanson2017julia}, leveraging its potent SciML Ecosystem~\citep{rackauckas2017differentialequations} which enables us to utilize a wide spectrum of differential equation classes (including SDEs, DDEs, DAEs) and a diverse suite of advanced solvers and sensitivity algorithms~\citep{rackauckas2019diffeqflux, 9622796}.

As reported in the literature, the identification of nonlinear dynamical systems, and in particular the gradient-descend based training of neural DE models, can be often challenging due to their highly complex loss landscapes, leading to poor local minima training stagnation~\citep{ribeiro2020smoothness, turan2021multiple}. We propose an enhancement to the original GOKU-net architecture which adds attention mechanisms to the main part of the model that infers the parameters of the differential equations. Moreover, to overcome the inherent difficulties of training, we developed a novel strategy to train the GOKU-net based on the multiple shooting technique~\citep{bock1984multiple, ribeiro2020smoothness, turan2021multiple} in the latent space. We have evaluated our enhanced model and training strategy on simulated data from a network of stochastic oscillators, specifically Stuart-Landau oscillators, as well as empirical brain data derived from resting state human functional Magnetic Resonance Imaging (fMRI). In both cases, the GOKU-net that fuses multiple shooting and attention, labeled \emph{GOKU-nets with Ubiquitous Inference} (GOKU-UI), outperformed both the base GOKU-net model and the baseline models in terms of reconstruction accuracy, forecasting capability, and data efficiency. We believe that GOKU-UI represents a promising step forward in Scientific Machine Learning, underscoring the rich possibilities that emerge when melding traditional scientific insights with contemporary machine learning techniques.

\section{Methods}
\subsection{Basic GOKU-nets}
\begin{figure}[t]
    \centering
    \includegraphics[trim={2cm 6cm 2cm 12cm},clip, width=1\linewidth]{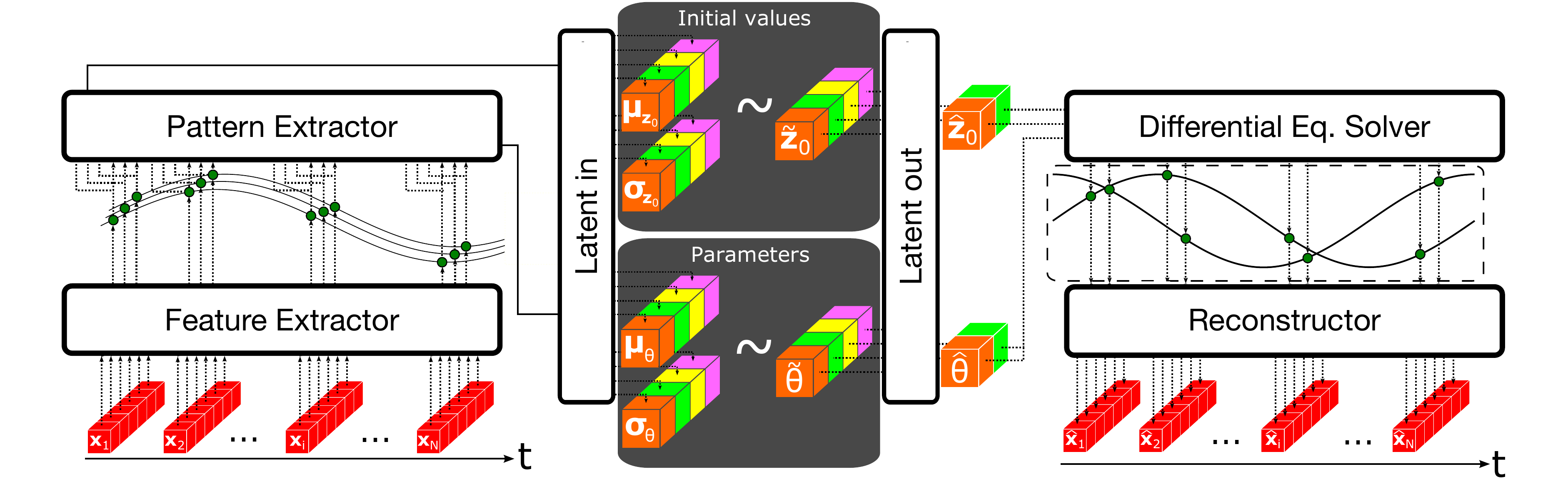}
    \caption{Schematic representation of a general Latent Differential Equation model. First, each time frame of the input data ${x_i}$ is independently processed by a \emph{Feature Extractor}, then the whole sequence goes through a \emph{Pattern Extractor}, which learns the mean and variance of the initial conditions, and possibly also the parameters, of the differential equation that will be next integrated. Finally, the solution goes through a last transformation performed by a \emph{Reconstructor} in order to go back to the input space.}
    \label{fig:LatentDiffeq_schema}
\end{figure}

GOKU-nets could be thought as a particular case of a more general model class that we call a Latent Differential Equation model (Latent DE), which schema is displayed in Figure \ref{fig:LatentDiffeq_schema}. Initially, each temporal frame of the input data ${x_i}$ is independently processed by a \emph{Feature Extractor}, usually reducing its dimensionality. Following this, the entire sequence is subjected to a \emph{Pattern Extractor}, which aims to learn the distribution of the initial conditions and possibly of the parameters for the differential equation that will be subsequently integrated. Lastly, the solution undergoes a final transformation via a \emph{Reconstructor}, going back to the original input space. The original model is trained as a standard VAE, maximizing the evidence lower bound (ELBO)~\citep{kingma2013auto}.

In the case of the Latent ODEs proposed by~\citet{chen_2018}, an RNN is used for the Pattern Extractor, a fully connected NN for the Reconstructor, and the differential equation is parametrized with another NN. This means that in this model the specific form of the differential equation is not provided beforehand but is learned simultaneously during training. The Feature Extractor and the intermediate layers, \emph{Latent in} and \emph{Latent out}, are not present so they could be considered as identity operations. On the other hand, the GOKU-net model proposed by~\citet{Linial_2021} has a ResNet with fully connected NNs as the Feature Extractor, while in the Pattern Extractor, an RNN is used to learn the initial conditions and a bidirectional LSTM for the ODE parameters. In this case the differential equation is explicitly predefined, allowing to incorporate some prior knowledge about the dynamical nature of the system under consideration. The Latent in and out layers are fully connected NN and finally, the Reconstructor is another ResNet, similar to the initial one. In the next section, some enhancements to the original GOKU-net model are proposed.

\subsection{GOKU-UI}
\subsubsection{Attention mechanism}
The first modification is the addition of a basic attention mechanism~\citep{vaswani2017attention} to the Pattern Extractor, specifically in the part associated with the learning of the parameters of the differential equation. Namely, instead of keeping the last element of the bidirectional LSTM (BiLSTM) used in the original GOKU-net model, all of their sequential outputs pass through a dense layer with softmax activation to calculate the attentional scores that would weight the sum of all the BiLSTM outputs in order to obtain its final output.

\subsubsection{Multiple Shooting}
When training Neural DE models, gradients have to be calculated through differential equations with respect to its initial conditions and parameters, by means of some sensitivity algorithm~\citep{ma2021comparison}. This tends to produce highly complex loss landscapes~\citep{ribeiro2020smoothness, metz2021gradients}. The work of ~\citet{turan2021multiple} demonstrates that training Neural ODEs even on very simple oscillatory data could be problematic, showing that the outcome may result in a trajectory similar to a moving average of the original data, thus failing to capture responses of higher frequency. They proposed a solution based on the \emph{multiple shooting} methods, which are widely used in Optimal Control~\citep{bock1984multiple, diehl2006fast} and Systems Identification~\citep{baake1992fitting, ribeiro2020smoothness} to alleviate the problem of high sensitivity to initial conditions and lower the probability of getting trapped at local minima with very poor performance. The basic idea of multiple shooting is to partition the complete time span over which the differential equation would be integrated into smaller time windows, for each of which the initial conditions are inferred in parallel. Afterwards, the multiple segments are joined and a continuity constraint is imposed during optimization, in our case, through the penalty method, which in practice simply consists of adding a regularization term to the loss function.

However, applying the multiple shooting method to GOKU-nets is not straightforward. Firstly, in most cases that use this method, such as in~\citet{turan2021multiple}, the differential equations are typically directly modeling the observable data, having direct access to the true initial conditions for each window. In the case of GOKU-nets, the dynamics modeled by differential equations occur in the latent space, which is being learned simultaneously; as a result, such true initial conditions are not available. Secondly, it is necessary to determine how the method will behave in relation to the parameters of the differential equation, which in the case of Neural ODEs are implicitly learned as part of their parameterization through the neural network.

Our proposal for extending the multiple shooting method to GOKU-nets is as follows. After passing through the Feature Extractor, we divide the temporal interval in the latent space in such a way that the Pattern Extractor generates in parallel different initial conditions for each temporal window, but provides a single set of parameters for the differential equations that will be shared by all windows. By this strategy, we maintain the potential benefits inherent to the multiple shooting method while leveraging the information available in a wider temporal range for the task of parameter inference, which is generally more challenging than estimating initial conditions. As mentioned before, we do not have access to target true initial conditions, however, what we can strive to achieve is the continuity of trajectories across different windows. To this end, these intervals are defined by overlapping the last temporal point of each window with the first one of the following and the goal is to minimize the distance between these points. Specifically, we employ regularization in the cost function when training the model, quadratically penalizing the discrepancy in the latent space of the overlapping points, that is, between the initial condition of each window and the end point of its preceding segment.

Our experiments indicated that non-variational GOKU-nets models outperform their variational counterparts significantly (see Supplementary Information \ref{sec:supplemental-further}). Therefore, we used non-variational GOKU-nets for all the remaining results in this work. Specifically, instead of sampling from normal distributions in the latent space as shown in Figure \ref{fig:LatentDiffeq_schema}, we directly take the mean values $\mu_{z_0}$ and $\mu_{\theta}$. As a result, the cost function associated with these models does not include the KL divergence term associated with the ELBO, but it does retain the reconstruction term, which is calculated as the mean squared error between the model's output and the input, normalized by the mean absolute value of the input. Furthermore, when using multiple shooting training, the continuity regularization described in the previous paragraph is included.

\subsection{Experiments}
In the next sections, we evaluate our proposed attention and multiple shooting enhancements through two highly challenging cases; one on synthetic data based on a network of stochastic oscillators known as Stuart-Landau oscillators and the other on empirical human brain data.

We compare the reconstruction and forecast performance of different variations of the GOKU-model (basic or with attention) trained in the original single shooting fashion or with the proposed multiple shooting method, as well as some baseline models: LSTM, Latent ODE, and a naïve model. For a fair comparison, both the LSTM and Latent ODE models are constructed maintaining the same GOKU-net general architecture and changing only the differential equation layer. Specifically, the Feature Extractor, Pattern Extractor, Latent In, Latent Out, and Reconstructor layers (see Figure \ref{fig:LatentDiffeq_schema}) maintain the same architecture and hyperparameters. However, the differential equation layer is substituted with a Neural ODE for the Latent ODE model, and in the other case, it is replaced by a recursively executed LSTM. The latent state dimensionality and size of the NN parameterizing the differential equation inside the Latent ODE, as well as the number of neurons in the LSTM were selected to match the total number of parameters of their contending GOKU-UI (with attention and multiple shooting). The naïve predictors, both for the reconstruction and forecast task, are simply constant predictions with the values of time-averaged inputs.

All models were trained under identical conditions and following the same procedure, with the aim of minimizing reconstruction error. In all instances, the input sequences to the model consisted of 46 time steps. During each training epoch, a random interval of this length was selected within the training data available for each sample in every batch of 64 samples. In the case of the multi-shooting training, the 46-time step sequence was further partitioned in the latent space into five windows, each comprising 10 time steps, with the last point of one window overlapping with the first point of the subsequent window. Under these circumstances, the loss function was augmented by the sum of squared differences between the endpoints of each segment. This sum was normalized by the number of junctions and multiplied by a regularization coefficient to impose a continuity constraint among the different segments. In the results presented here, the regularization coefficient was set to 2. Comprehensive details of the training process, as well as the specific architecture of the models and the hyperparameters used, can be found in the Supplementary Information.

\subsubsection{Simulated data}

Stuart-Landau (SL) oscillators, representing the normal form of a supercritical Hopf bifurcation, serve as a fundamental archetype of mathematical models and are extensively used across diverse scientific disciplines to study self-sustained oscillations~\citep{kuramoto1984chemical, kuznetsov1998elements}. The SL oscillator is often described by the complex-valued differential equation 
\begin{eqnarray}
\label{eq:complex-stuart-landau}
\dot{z} = z(a + i\omega) - z |z|^2
\end{eqnarray}
where $z = \rho e^{i\theta} = x + iy$, $a$ is the bifurcation parameter, and $\omega$ is intrinsic frequency of the oscillator. The parameter $a$ represents the linear growth rate of the system. When $a$ is positive, the amplitude of the oscillation increases, and when a is negative, the amplitude of the oscillation decreases. At $a=0$, a Hopf bifurcation occurs, and the system transitions from a stable fixed point to limit cycle oscillations (or vice versa). Despite its apparent simplicity, the SL model can exhibit a wide range of behaviors, including limit cycles, chaotic oscillations, and quasiperiodicity, making it a versatile tool in the study of nonlinear dynamics and a good candidate for evaluating the capabilities of the GOKU-net models.

In particular, we generate the simulated data with a network of coupled stochastic Stuart-Landau oscillators that has been widely used to model brain dynamics for resting state fMRI \citep{jobst2017increased, deco2017dynamics, donnelly2019reliable, ipina2020modeling}, which will also be used in the empirical evaluation on brain data, described in the next section. The dynamics of the $i$-th node within a network of N oscillators is given by the following equation:
\begin{eqnarray}
\label{eq:stuart-landau_network}
\dot{x}_{j} = Re (\dot{z}_{j}) = [a_j - x^2_j - y^2_j]x_j - \omega_j y_j + G \sum_{i=1}^{N} C_{ij} (x_i - x_j) + \beta \eta_j (t) \nonumber \\
\dot{y}_{j} = Im(\dot{z}_{j}) = [a_j - x^2_j - y^2_j]y_j - \omega_j x_j + G \sum_{i=1}^{N} C_{ij} (y_i - y_j) + \beta \eta_j (t)
\end{eqnarray}

where $C_{ij}$ is a connectivity matrix between all nodes in the network, $G$ is a global coupling factor, while $\eta_j$ represents additive Gaussian noise. Note that independent bifurcation parameters $a_j$ and frequencies $\omega_j$ are used for each node.

During the construction of our dataset, we perform a dimensionality augmentation on the network of oscillators, which are utilized as latent dynamics. Specifically, we apply a random linear transformation, $f: \mathbb{R}^{2N} \rightarrow \mathbb{R}^{D}$, to the latent trajectories of each sample, where the dimension $D$ is much larger than $2N$. Each sample corresponds to a unique random set of initial conditions and parameters for the $N$ coupled oscillators. All the synthetic data experiments were performed using $N = 3$ stochastic oscillators with a high dimension $D = 784$. All details of the implementation and hyperparameters can be found in the Supplementary Information, and the codes are accessible in the GitHub repository\footnote{The link will be provided upon acceptance.}.

\subsubsection{Empiric data}
Next, in an effort to evaluate both our proposed models in a challenging empirical data set and to provide an example of the advantages of incorporating prior scientific insights into agnostic AI models, we focus on one of the most complex systems in nature: the human brain.

We used the resting state fMRI data from 153 subjects, sourced from the Track-On HD study~\citep{kloppel2015compensation}. The data was pre-processed as described in~\citet{polosecki2020resting}, followed by a 20-component Canonical ICA~\citep{varoquaux2010group}. Of the original 20 components, 9 were identified as artifacts and therefore eliminated, leaving 11 components for further analysis. The contribution of each subject was captured in two separate visits, yielding an aggregate of 306 data samples. These samples, with 160 time points each, were acquired at a temporal resolution of 3 seconds. For the scope of our investigation, we set aside roughly 20\% of the data samples (n=60) for testing, while ensuring a balanced representation of the sex, condition, and measurement site. The remaining data samples (n = 246) were assigned for training and validation. More specifically, the first 114 time points from each of these samples were used for model training, with the remainder reserved for validation and early training termination.

Our proposed GOKU-UI model was trained using 20 Stuart-Landau stochastic oscillators (Eq. \ref{eq:stuart-landau_network}) in the latent space. A comparison of the model's performance using various numbers of oscillators and a thorough description of the training procedure are presented in the Supplementary Information section.

\section{Results}

\label{sec:results}
\begin{figure}[!bp]
    \centering
    \begin{subfigure}[b]{0.49\textwidth}
        \centering
        \subcaption{}
        \includegraphics[width=\textwidth]{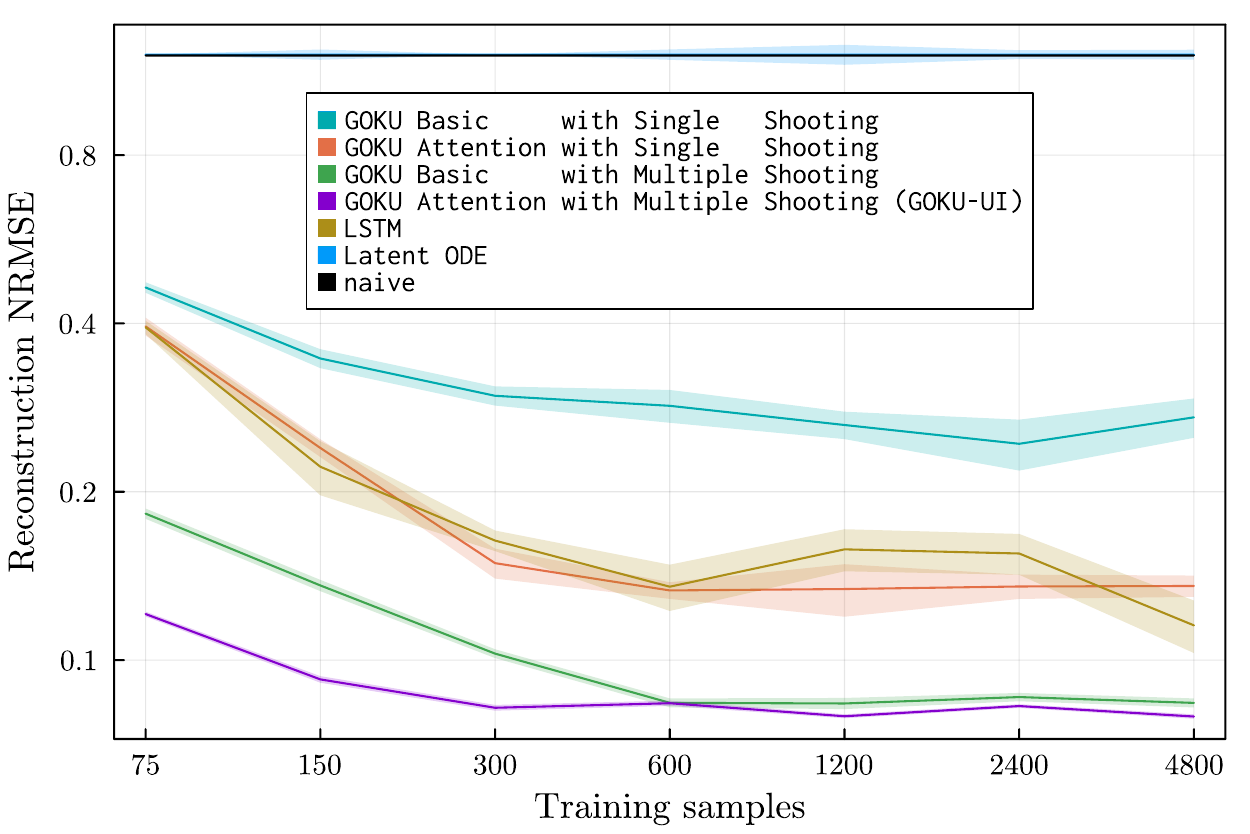}
        \label{fig:Hopf_data_scaling}
    \end{subfigure}
    \hfill
    \begin{subfigure}[b]{0.49\textwidth}
        \centering
        \subcaption{}
        \includegraphics[width=\textwidth]{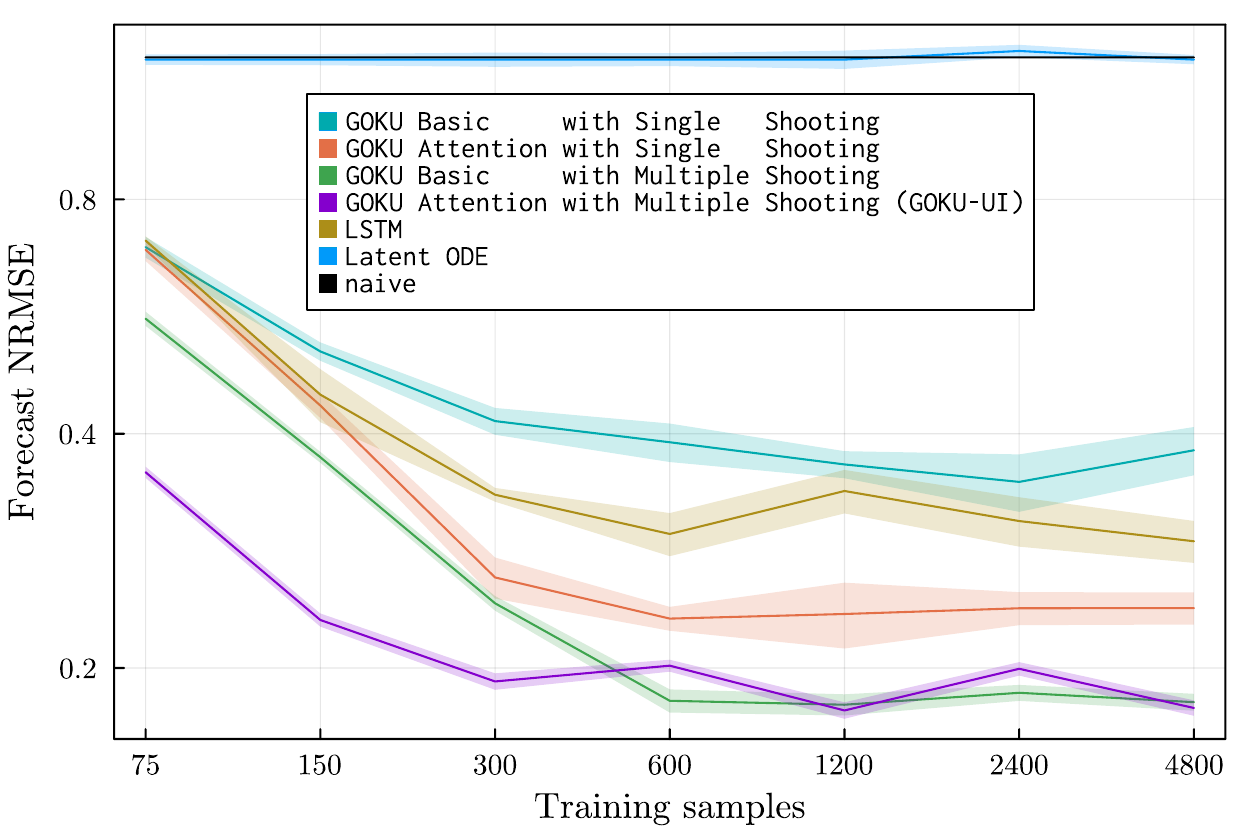}
        \label{fig:hopf_data_scaling_forecast}
    \end{subfigure}
    \caption{Comparison of the reconstruction ({\subref{fig:Hopf_data_scaling}}) and forecast ({\subref{fig:hopf_data_scaling_forecast}}) performances on the synthetic Stuart-Landau dataset measured by the median normalized RMSE while increasing the number of samples in the training set. The averages were taken with respect to the input dimensions and time span, while the shaded regions correspond to standard errors derived from various training runs with multiple random seeds. The forecast was evaluated on a 20 time-steps horizon.}
    \label{fig:hopf}
\end{figure}

We evaluated the performance of four GOKU-net variants across both reconstruction and forecast tasks. These variants encompass single and multiple shooting methods, either with or without the use of the attention mechanism. Comparisons were made with three baseline models, namely LSTM, Latent ODEs, and a naïve predictor.

All models were exclusively trained on the reconstruction task and their forecasts were generated during the evaluation stage. The prediction error was quantified using the normalized root mean square error (NRMSE) between the target ground truth and either the reconstruction or the forecast.

\subsection{Simulated data}
Figure \ref{fig:Hopf_data_scaling} shows the performance of the different models in the reconstruction task on a synthetic dataset generated with the latent dynamics of three stochastic Stuart-Landau oscillators. In this context, the GOKU-net variants employing the multiple shooting method demonstrated significantly lower errors compared to other models. Notably, GOKU-nets that utilized the attention mechanism consistently showed better outcomes relative to the basic variant. The single shooting variant particularly benefited from the attention mechanism.
In the multiple shooting scenario, the performance boost due to attention was pronounced when using fewer training samples. However, when the number of training samples exceeded 600, the performance of both models converged to much closer values. However, GOKU-UI, which incorporates both the attention mechanism and multiple shooting, was still significantly the best-performing model. Wilcoxon signed rank tests, comparing GOKU-UI with each of the other models, for each number of training samples independently, yielded all p-values~<~0.02 after Holm correction.

In contrast, the Latent ODEs, whose training runs consistently became trapped in local minima associated with the mean of the time series, markedly underperformed and exhibited errors akin to those of the naïve predictor. On the other hand, the LSTMs outperformed the basic GOKU-nets trained with the single shooting method. GOKU-UI, when trained on a mere 300 samples, yielded a reconstruction performance comparable to that achieved with 4800 samples. Moreover, when trained on just 75 unique samples, GOKU-UI outperformed all other single shooting models, even those trained on a dataset 64 times larger.

\begin{figure}[!b]
    \centering
    \begin{subfigure}[b]{0.49\textwidth}
        \centering
        \subcaption{}
        \includegraphics[width=\textwidth]{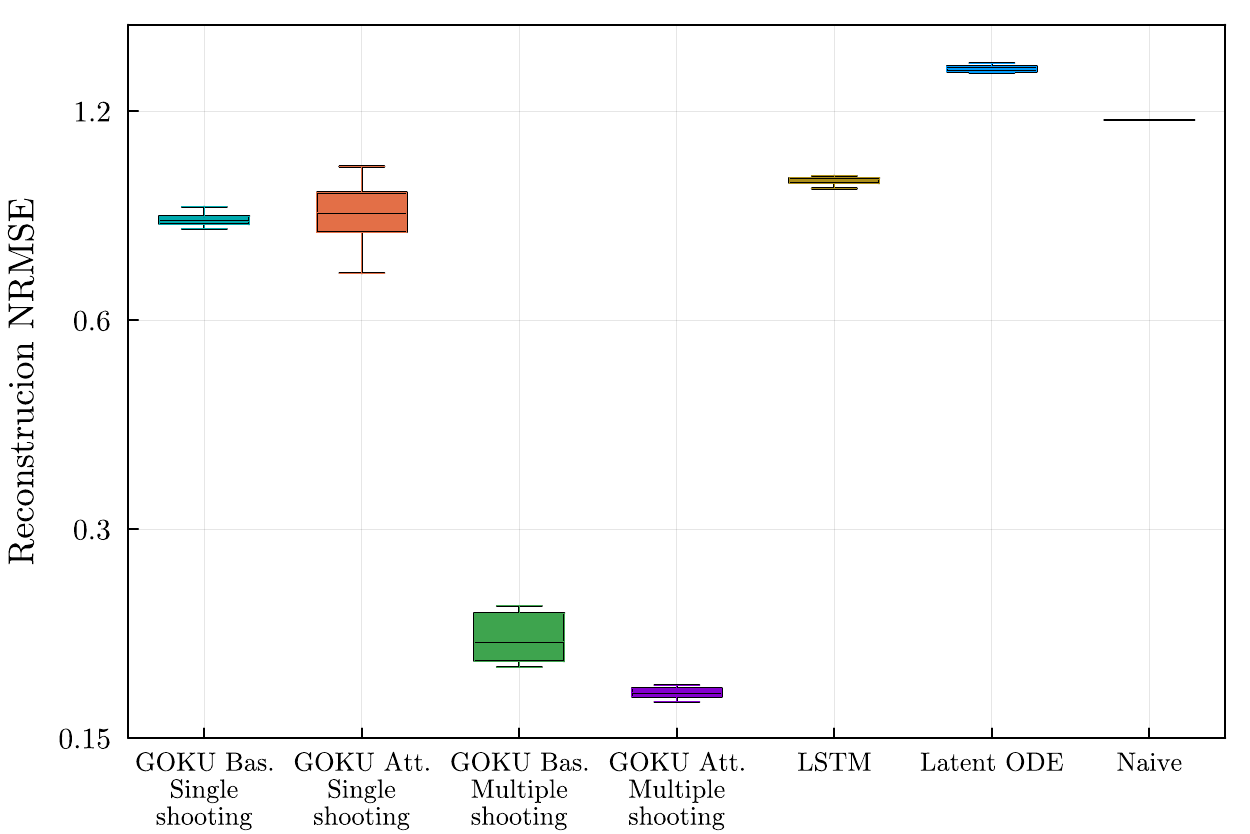}
        \label{fig:fMRI_reconstruction_baselines}
    \end{subfigure}
    \hfill
    \begin{subfigure}[b]{0.49\textwidth}
        \centering
        \subcaption{}
        \includegraphics[width=\textwidth]{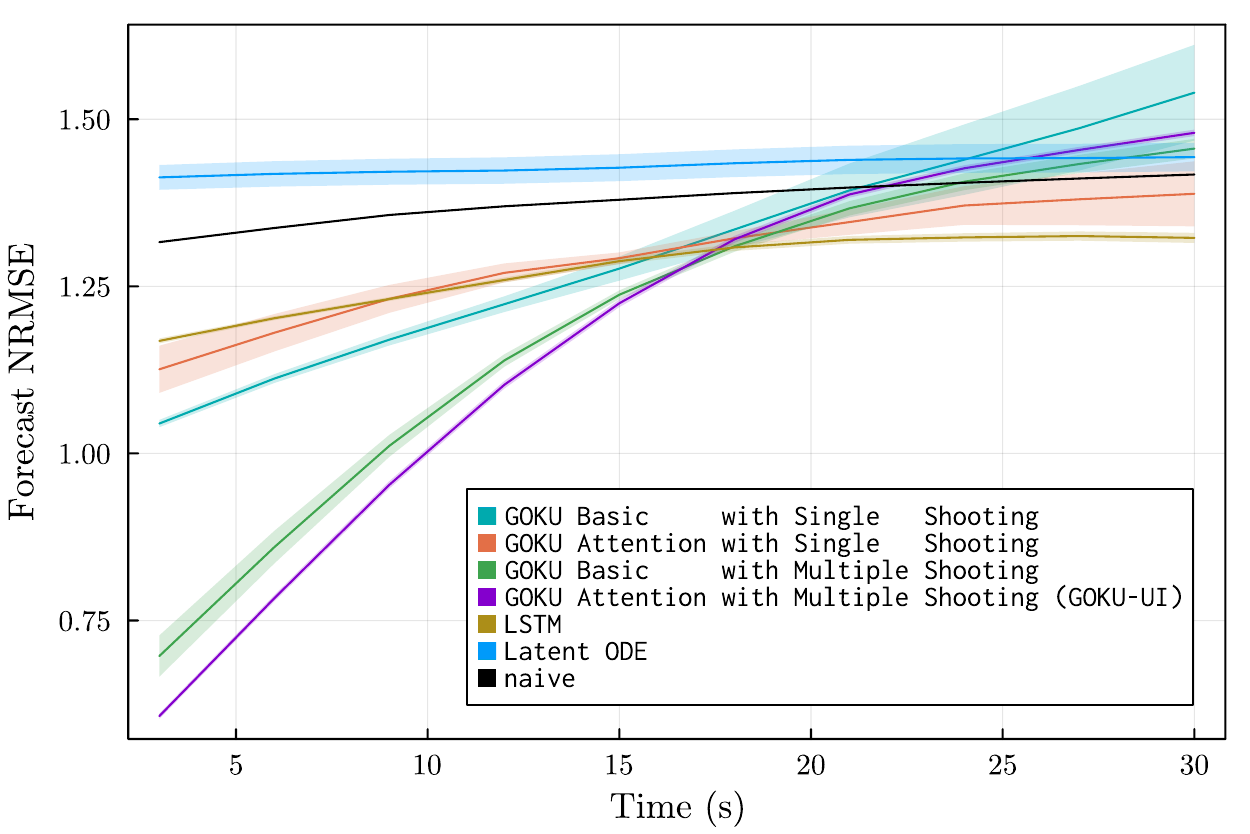}
        \label{fig:fMRI_forecast_baselines}
    \end{subfigure}
    \caption{Comparison of the reconstruction ({\subref{fig:fMRI_reconstruction_baselines}}) and forecast ({\subref{fig:fMRI_forecast_baselines}}) performances on testing fMRI data, measured by the normalized RMSE. The averages were taken with respect to the input dimensions and time span, while the shaded regions correspond to standard errors derived from various training runs with multiple random seeds.}
    \label{fig:fMRI}
\end{figure}

When forecasting 20 temporal points beyond the reconstruction limit, we observed a similar trend, as depicted in Figure \ref{fig:hopf_data_scaling_forecast}. However, in this task, GOKU-nets trained with single shooting and incorporating the attention mechanism surpassed the performance of the LSTMs. GOKU-nets utilizing the multiple shooting method continued to exhibit superior overall performance. GOKU-UI significantly outperformed other models up to 150 training samples (all p-values~<~0.02, according to Wilcoxon signed-rank tests conducted independently for each sample size, after Holm correction), beyond which its performance was statistically indistinguishable from that of the basic GOKU model with multiple shooting (p-values~>~0.05, Wilcoxon signed-rank tests, Holm corrected).

\subsection{Empirical data}
In a separate analysis, we trained GOKU-net models on human brain recordings, employing 20 coupled stochastic Stuart-Landau oscillators to govern the dynamics in their latent space.  These models were trained alongside baseline models on 246 samples of 11-component ICA time series obtained from fMRI data. As shown in Figure \ref{fig:fMRI_reconstruction_baselines}, unlike the simulated data scenario, the addition of the attention mechanism did not improve the reconstruction performance of the single shooting GOKU-net models. However, the multiple shooting training method again resulted in a significant improvement factor (p~<~0.001, Wilcoxon signed-rank test). Remarkably, the GOKU-UI model, which integrates both multiple shooting and the attention mechanism, achieved a median reconstruction error that was five times lower than that of single shooting baseline models. Specifically, GOKU-UI also exhibited a significantly lower reconstruction NRMSE compared to the basic GOKU model that employed multiple shooting (p~<~0.04, Wilcoxon signed-rank test). For a qualitative performance comparison between GOKU-UI and the original GOKU-net, refer to representative reconstruction plots in the Supplementary Information \ref{sec:rec_plots}.

Models trained exclusively for reconstruction were also assessed in a forecast task, where they were tasked with predicting time series evolution immediately beyond the reconstruction limit. Figure \ref{fig:fMRI_forecast_baselines} illustrates the performance trends of the different models as the forecast horizon is extended. In this scenario, GOKU-UI outperformed other models, achieving lower forecast errors for up to 15 seconds of brain activity, thus further confirming its improved performance in both reconstruction and forecast tasks.

\section{Discussion and Limitations}

Our research focused on two key enhancements to the baseline GOKU-net model: the addition of a basic attention mechanism and the implementation of a multiple shooting method. Independently, each of these modifications tended to improve the model's performance in both reconstruction and forecast tasks. The use of multiple shooting yielded the most substantial improvement. In particular, GOKU-UI, a composite of both enhancements, exhibited the best overall performance. During the evaluation on the synthetic dataset, GOKU-UI demonstrated remarkable efficiency with respect to the training data. With just 300 training samples, the model achieved performance comparable to when it used 16 times as many samples. Additionally, with a mere 150 training samples, it outperformed all other single shooting baseline models, even when those had employed training sets 32 times larger (Figure \ref{fig:hopf}). In the case of the empirical dataset, compared to the original GOKU-nets, GOKU-UI managed to accurately capture the complex brain dynamics, achieving a five-fold reduction in reconstruction error, and surpassing all baseline methods in the forecast task up to 15 seconds beyond the reconstruction limit (Figure \ref{fig:fMRI}).

Furthermore, we implemented GOKU-nets in the Julia Programming Language, which broadened the capabilities of the model. It helped overcome the initial limitation to Ordinary Differential Equations (ODEs) and facilitated the use of a wider range of differential equation classes (e.g., SDEs), as well as alternative advanced solvers and sensitivity algorithms. The unique SciML ecosystem of the Julia Language proved to be a potent and effective tool for research at the intersection of dynamical systems and machine learning.

Although the Stuart-Landau model has appeared in prior literature modeling brain dynamics~\citep{deco2017dynamics, donnelly2019reliable, ipina2020modeling}, it has generally been employed with fixed coupling between oscillators. In these empirical studies, structural connectivity estimates were derived from Diffusion Tensor Imaging (DTI). Other parameters were adjusted to maximize some goodness-of-fit criterion between the simulated and empirical averaged functional connectivities. In contrast to those studies, the methodology employed in the current work targets the minimization of residuals between the simulated and empirical time series themselves.
On the other hand, to our knowledge, this is the first instance of employing Stuart-Landau oscillators to model latent brain data dynamics while simultaneously inferring oscillator connectivity and learning the nonlinear transformation into the latent space. This approach differs from previous works, such as~\citet{abrevaya2021learning}, which fitted the more general van der Pol oscillators to the latent representation of brain data. Specifically, their learning process involved two separate procedures: encoding in the latent space and parameter estimation of the differential equations. In contrast, our GOKU-UI model integrates these processes into a single end-to-end training regime. As discussed in~\cite{ramezanian2022generative}, this approach provides the unique advantage of integrating the known dynamics that govern the system, thus facilitating the interpretability and potential applicability with a smaller training set, while still maintaining the flexibility to learn nonlinear dependencies. 

We have demonstrated these capabilities by training the GOKU-UI on fMRI data and encoding whole-brain dynamics into a latent representation. This representation's temporal evolution is effectively modeled by a low-dimensional, interpretable dynamical system, which can yield profound insights into brain functionality, such as the inference of functional connectivity. Beyond theoretical understanding, the model also holds potential for applied usage, including the classification of mental states or psychiatric conditions. These applications could either leverage the parameters of the differential equations or exploit higher-level features of the latent system, such as its attractor topology.

Despite its versatility as a tool in the SciML toolbox, GOKU-UI's main advantage may sometimes also be its primary limitation: unlike traditional, more agnostic machine learning models, GOKU-UI requires a preliminary differential equation model hypothesized to govern the data's intrinsic temporal dynamics. This requirement may be challenging to meet in many cases. For example, with latent ODEs, one can bypass this task, allowing another neural network to learn the differential equation. However, the significant complexity of the system under investigation, as evidenced by our experiments, could potentially hinder the efficacy of this method.

To the uninitiated in the field of dynamical systems, the process of proposing a specific differential equation to model the data's intrinsic but not immediately evident dynamics might seem like a guessing game. However, this approach has been the successful foundation of the field of physics since the era of Newton. The application of GOKU-UI to a new problem might not be as straightforward as a general-purpose black-box neural network model. Still, when guided by the vast theory of dynamical systems, it is not only possible, but potentially highly fruitful.

\section{Acknowledgments}

Guillaume Dumas was supported by the Institute for Data Valorization, Montreal (IVADO; CF00137433) Professor Startup \& Operational Funds, the Fonds de la Recherche en Santé du Québec (FRQ; 295289; 295291) Junior 1 salary award, the Natural Sciences and Engineering Research Council of Canada (NSERC; DGECR-2023-00089), and the Azrieli Global Scholars Fellowship from the Canadian Institute for Advanced Research (CIFAR) in the Brain, Mind, \& Consciousness program. Irina Rish and Mahta Ramezanian Panahi acknowledge the support from the Canada CIFAR AI Chair Program, the Canada Excellence Research Chairs (CERC) program. Furthermore, Mahta Ramezanian Panahi acknowledges the UNIQUE Center support. Finally, the work of Silvina Ponce Dawson and Germán Abrevaya was funded by UBA (UBACyT 20020170100482BA) and ANPCyT (PICT-2018-02026, PICT-2021-I-A-00128, PICT-2021-III-A-00091).
The computational resources used in this work were provided (in part) by the HPC center DIRAC, funded by Instituto de Fisica de Buenos Aires (UBA-CONICET) and part of the SNCAD-MinCyT initiative, Argentina.

\bibliography{main}

\begin{thebibliography}{57}
\providecommand{\natexlab}[1]{#1}
\providecommand{\url}[1]{\texttt{#1}}
\expandafter\ifx\csname urlstyle\endcsname\relax
  \providecommand{\doi}[1]{doi: #1}\else
  \providecommand{\doi}{doi: \begingroup \urlstyle{rm}\Url}\fi

\bibitem[Abrevaya et~al.(2021)Abrevaya, Dumas, Aravkin, Zheng, Gagnon-Audet,
  Kozloski, Polosecki, Lajoie, Cox, Dawson, et~al.]{abrevaya2021learning}
Abrevaya, G., Dumas, G., Aravkin, A.~Y., Zheng, P., Gagnon-Audet, J.-C.,
  Kozloski, J., Polosecki, P., Lajoie, G., Cox, D., Dawson, S.~P., et~al.
\newblock Learning brain dynamics with coupled low-dimensional nonlinear
  oscillators and deep recurrent networks.
\newblock \emph{Neural Computation}, 33\penalty0 (8):\penalty0 2087--2127,
  2021.

\bibitem[Baake et~al.(1992)Baake, Baake, Bock, and Briggs]{baake1992fitting}
Baake, E., Baake, M., Bock, H., and Briggs, K.
\newblock Fitting ordinary differential equations to chaotic data.
\newblock \emph{Physical Review A}, 45\penalty0 (8):\penalty0 5524, 1992.

\bibitem[Baker et~al.(2019)Baker, Alexander, Bremer, Hagberg, Kevrekidis, Najm,
  Parashar, Patra, Sethian, Wild, Willcox, and Lee]{osti_1478744}
Baker, N., Alexander, F., Bremer, T., Hagberg, A., Kevrekidis, Y., Najm, H.,
  Parashar, M., Patra, A., Sethian, J., Wild, S., Willcox, K., and Lee, S.
\newblock Workshop report on basic research needs for scientific machine
  learning: Core technologies for artificial intelligence, 2 2019.
\newblock URL \url{https://www.osti.gov/biblio/1478744}.

\bibitem[Bezanson et~al.(2017)Bezanson, Edelman, Karpinski, and
  Shah]{bezanson2017julia}
Bezanson, J., Edelman, A., Karpinski, S., and Shah, V.~B.
\newblock Julia: A fresh approach to numerical computing.
\newblock \emph{SIAM review}, 59\penalty0 (1):\penalty0 65--98, 2017.

\bibitem[Bock \& Plitt(1984)Bock and Plitt]{bock1984multiple}
Bock, H.~G. and Plitt, K.-J.
\newblock A multiple shooting algorithm for direct solution of optimal control
  problems.
\newblock \emph{IFAC Proceedings Volumes}, 17\penalty0 (2):\penalty0
  1603--1608, 1984.

\bibitem[Brunton et~al.(2016)Brunton, Proctor, and Kutz]{brunton_2016}
Brunton, S.~L., Proctor, J.~L., and Kutz, J.~N.
\newblock Discovering governing equations from data by sparse identification of
  nonlinear dynamical systems.
\newblock \emph{Proceedings of the National Academy of Sciences}, 113\penalty0
  (15):\penalty0 3932--3937, 2016.
\newblock \doi{10.1073/pnas.1517384113}.
\newblock URL \url{https://www.pnas.org/doi/abs/10.1073/pnas.1517384113}.

\bibitem[Chang et~al.(2018)Chang, Meng, Haber, Ruthotto, Begert, and
  Holtham]{chang2018reversible}
Chang, B., Meng, L., Haber, E., Ruthotto, L., Begert, D., and Holtham, E.
\newblock Reversible architectures for arbitrarily deep residual neural
  networks.
\newblock In \emph{Proceedings of the AAAI conference on artificial
  intelligence}, volume~32, 2018.

\bibitem[Chen et~al.(2018)Chen, Rubanova, Bettencourt, and Duvenaud]{chen_2018}
Chen, R. T.~Q., Rubanova, Y., Bettencourt, J., and Duvenaud, D.
\newblock Neural ordinary differential equations, 2018.
\newblock URL \url{https://arxiv.org/abs/1806.07366}.

\bibitem[Datseris et~al.(2020)Datseris, Isensee, Pech, and Gál]{Datseris2020}
Datseris, G., Isensee, J., Pech, S., and Gál, T.
\newblock Drwatson: the perfect sidekick for your scientific inquiries.
\newblock \emph{Journal of Open Source Software}, 5\penalty0 (54):\penalty0
  2673, 2020.
\newblock \doi{10.21105/joss.02673}.
\newblock URL \url{https://doi.org/10.21105/joss.02673}.

\bibitem[Deco et~al.(2017)Deco, Kringelbach, Jirsa, and
  Ritter]{deco2017dynamics}
Deco, G., Kringelbach, M.~L., Jirsa, V.~K., and Ritter, P.
\newblock The dynamics of resting fluctuations in the brain: metastability and
  its dynamical cortical core.
\newblock \emph{Scientific reports}, 7\penalty0 (1):\penalty0 3095, 2017.

\bibitem[Diehl et~al.(2006)Diehl, Bock, Diedam, and Wieber]{diehl2006fast}
Diehl, M., Bock, H.~G., Diedam, H., and Wieber, P.-B.
\newblock Fast direct multiple shooting algorithms for optimal robot control.
\newblock \emph{Fast motions in biomechanics and robotics: optimization and
  feedback control}, pp.\  65--93, 2006.

\bibitem[Donnelly-Kehoe et~al.(2019)Donnelly-Kehoe, Saenger, Lisofsky,
  K{\"u}hn, Kringelbach, Schwarzbach, Lindenberger, and
  Deco]{donnelly2019reliable}
Donnelly-Kehoe, P., Saenger, V.~M., Lisofsky, N., K{\"u}hn, S., Kringelbach,
  M.~L., Schwarzbach, J., Lindenberger, U., and Deco, G.
\newblock Reliable local dynamics in the brain across sessions are revealed by
  whole-brain modeling of resting state activity.
\newblock \emph{Human Brain Mapping}, 40\penalty0 (10):\penalty0 2967--2980,
  2019.

\bibitem[Foias et~al.(1988)Foias, Sell, and Temam]{FOIAS1988309}
Foias, C., Sell, G.~R., and Temam, R.
\newblock Inertial manifolds for nonlinear evolutionary equations.
\newblock \emph{Journal of Differential Equations}, 73\penalty0 (2):\penalty0
  309--353, 1988.
\newblock ISSN 0022-0396.
\newblock \doi{https://doi.org/10.1016/0022-0396(88)90110-6}.
\newblock URL
  \url{https://www.sciencedirect.com/science/article/pii/0022039688901106}.

\bibitem[Goodfellow et~al.(2014)Goodfellow, Pouget-Abadie, Mirza, Xu,
  Warde-Farley, Ozair, Courville, and Bengio]{GAN_NIPS2014}
Goodfellow, I., Pouget-Abadie, J., Mirza, M., Xu, B., Warde-Farley, D., Ozair,
  S., Courville, A., and Bengio, Y.
\newblock Generative adversarial nets.
\newblock In Ghahramani, Z., Welling, M., Cortes, C., Lawrence, N., and
  Weinberger, K. (eds.), \emph{Advances in Neural Information Processing
  Systems}, volume~27. Curran Associates, Inc., 2014.
\newblock URL
  \url{https://proceedings.neurips.cc/paper/2014/file/5ca3e9b122f61f8f06494c97b1afccf3-Paper.pdf}.

\bibitem[Guo \& Hesthaven(2019)Guo and Hesthaven]{GUO201975}
Guo, M. and Hesthaven, J.~S.
\newblock Data-driven reduced order modeling for time-dependent problems.
\newblock \emph{Computer Methods in Applied Mechanics and Engineering},
  345:\penalty0 75--99, 2019.
\newblock ISSN 0045-7825.
\newblock \doi{https://doi.org/10.1016/j.cma.2018.10.029}.
\newblock URL
  \url{https://www.sciencedirect.com/science/article/pii/S0045782518305334}.

\bibitem[Haber \& Ruthotto(2017)Haber and Ruthotto]{haber2017stable}
Haber, E. and Ruthotto, L.
\newblock Stable architectures for deep neural networks.
\newblock \emph{Inverse problems}, 34\penalty0 (1):\penalty0 014004, 2017.

\bibitem[He et~al.(2016)He, Zhang, Ren, and Sun]{he2016deep}
He, K., Zhang, X., Ren, S., and Sun, J.
\newblock Deep residual learning for image recognition.
\newblock In \emph{Proceedings of the IEEE conference on computer vision and
  pattern recognition}, pp.\  770--778, 2016.

\bibitem[Innes et~al.(2018)Innes, Saba, Fischer, Gandhi, Rudilosso, Joy,
  Karmali, Pal, and Shah]{Flux.jl-2018}
Innes, M., Saba, E., Fischer, K., Gandhi, D., Rudilosso, M.~C., Joy, N.~M.,
  Karmali, T., Pal, A., and Shah, V.
\newblock Fashionable modelling with flux.
\newblock \emph{CoRR}, abs/1811.01457, 2018.
\newblock URL \url{https://arxiv.org/abs/1811.01457}.

\bibitem[Ipi{\~n}a et~al.(2020)Ipi{\~n}a, Kehoe, Kringelbach, Laufs,
  Iba{\~n}ez, Deco, Perl, and Tagliazucchi]{ipina2020modeling}
Ipi{\~n}a, I.~P., Kehoe, P.~D., Kringelbach, M., Laufs, H., Iba{\~n}ez, A.,
  Deco, G., Perl, Y.~S., and Tagliazucchi, E.
\newblock Modeling regional changes in dynamic stability during sleep and
  wakefulness.
\newblock \emph{NeuroImage}, 215:\penalty0 116833, 2020.

\bibitem[Jobst et~al.(2017)Jobst, Hindriks, Laufs, Tagliazucchi, Hahn,
  Ponce-Alvarez, Stevner, Kringelbach, and Deco]{jobst2017increased}
Jobst, B.~M., Hindriks, R., Laufs, H., Tagliazucchi, E., Hahn, G.,
  Ponce-Alvarez, A., Stevner, A., Kringelbach, M.~L., and Deco, G.
\newblock Increased stability and breakdown of brain effective connectivity
  during slow-wave sleep: mechanistic insights from whole-brain computational
  modelling.
\newblock \emph{Scientific reports}, 7\penalty0 (1):\penalty0 1--16, 2017.

\bibitem[Kidger(2022)]{kidger2022neural}
Kidger, P.
\newblock On neural differential equations.
\newblock \emph{arXiv preprint arXiv:2202.02435}, 2022.

\bibitem[Kim et~al.(2019)Kim, Azevedo, Thuerey, Kim, Gross, and
  Solenthaler]{kim_2019}
Kim, B., Azevedo, V.~C., Thuerey, N., Kim, T., Gross, M., and Solenthaler, B.
\newblock Deep fluids: A generative network for parameterized fluid
  simulations.
\newblock \emph{Computer Graphics Forum}, 38\penalty0 (2):\penalty0 59--70,
  2019.
\newblock \doi{https://doi.org/10.1111/cgf.13619}.
\newblock URL \url{https://onlinelibrary.wiley.com/doi/abs/10.1111/cgf.13619}.

\bibitem[Kim et~al.(2022)Kim, Choi, Widemann, and Zohdi]{kim_2021}
Kim, Y., Choi, Y., Widemann, D., and Zohdi, T.
\newblock A fast and accurate physics-informed neural network reduced order
  model with shallow masked autoencoder.
\newblock \emph{Journal of Computational Physics}, 451:\penalty0 110841, 2022.
\newblock ISSN 0021-9991.
\newblock \doi{https://doi.org/10.1016/j.jcp.2021.110841}.
\newblock URL
  \url{https://www.sciencedirect.com/science/article/pii/S0021999121007361}.

\bibitem[Kingma \& Welling(2013)Kingma and Welling]{kingma2013auto}
Kingma, D.~P. and Welling, M.
\newblock Auto-encoding variational bayes.
\newblock \emph{arXiv preprint arXiv:1312.6114}, 2013.

\bibitem[Kl{\"o}ppel et~al.(2015)Kl{\"o}ppel, Gregory, Scheller, Minkova, Razi,
  Durr, Roos, Leavitt, Papoutsi, Landwehrmeyer,
  et~al.]{kloppel2015compensation}
Kl{\"o}ppel, S., Gregory, S., Scheller, E., Minkova, L., Razi, A., Durr, A.,
  Roos, R.~A., Leavitt, B.~R., Papoutsi, M., Landwehrmeyer, G.~B., et~al.
\newblock Compensation in preclinical huntington's disease: evidence from the
  track-on hd study.
\newblock \emph{EBioMedicine}, 2\penalty0 (10):\penalty0 1420--1429, 2015.

\bibitem[Kuramoto(1984)]{kuramoto1984chemical}
Kuramoto, Y.
\newblock \emph{Chemical Oscillations, Waves and Turbulence}.
\newblock Springer-Verlag, 1984.

\bibitem[Kuznetsov et~al.(1998)Kuznetsov, Kuznetsov, and
  Kuznetsov]{kuznetsov1998elements}
Kuznetsov, Y.~A., Kuznetsov, I.~A., and Kuznetsov, Y.
\newblock \emph{Elements of applied bifurcation theory}, volume 112.
\newblock Springer, 1998.

\bibitem[Linial et~al.(2021)Linial, Ravid, Eytan, and Shalit]{Linial_2021}
Linial, O., Ravid, N., Eytan, D., and Shalit, U.
\newblock Generative {ODE} modeling with known unknowns.
\newblock In \emph{Proceedings of the Conference on Health, Inference, and
  Learning}. {ACM}, apr 2021.
\newblock \doi{10.1145/3450439.3451866}.
\newblock URL \url{https://doi.org/10.1145%2F3450439.3451866}.

\bibitem[Lu et~al.(2018)Lu, Zhong, Li, and Dong]{lu2018beyond}
Lu, Y., Zhong, A., Li, Q., and Dong, B.
\newblock Beyond finite layer neural networks: Bridging deep architectures and
  numerical differential equations.
\newblock In \emph{International Conference on Machine Learning}, pp.\
  3276--3285. PMLR, 2018.

\bibitem[Lucia et~al.(2004)Lucia, Beran, and Silva]{LUCIA200451}
Lucia, D.~J., Beran, P.~S., and Silva, W.~A.
\newblock Reduced-order modeling: new approaches for computational physics.
\newblock \emph{Progress in Aerospace Sciences}, 40\penalty0 (1):\penalty0
  51--117, 2004.
\newblock ISSN 0376-0421.
\newblock \doi{https://doi.org/10.1016/j.paerosci.2003.12.001}.
\newblock URL
  \url{https://www.sciencedirect.com/science/article/pii/S0376042103001131}.

\bibitem[Ma et~al.(2021{\natexlab{a}})Ma, Dixit, Innes, Guo, and
  Rackauckas]{9622796}
Ma, Y., Dixit, V., Innes, M.~J., Guo, X., and Rackauckas, C.
\newblock A comparison of automatic differentiation and continuous sensitivity
  analysis for derivatives of differential equation solutions.
\newblock In \emph{2021 IEEE High Performance Extreme Computing Conference
  (HPEC)}, pp.\  1--9, 2021{\natexlab{a}}.
\newblock \doi{10.1109/HPEC49654.2021.9622796}.

\bibitem[Ma et~al.(2021{\natexlab{b}})Ma, Dixit, Innes, Guo, and
  Rackauckas]{ma2021comparison}
Ma, Y., Dixit, V., Innes, M.~J., Guo, X., and Rackauckas, C.
\newblock A comparison of automatic differentiation and continuous sensitivity
  analysis for derivatives of differential equation solutions.
\newblock In \emph{2021 IEEE High Performance Extreme Computing Conference
  (HPEC)}, pp.\  1--9. IEEE, 2021{\natexlab{b}}.

\bibitem[Ma et~al.(2021{\natexlab{c}})Ma, Gowda, Anantharaman, Laughman, Shah,
  and Rackauckas]{ma2021modelingtoolkit}
Ma, Y., Gowda, S., Anantharaman, R., Laughman, C., Shah, V., and Rackauckas, C.
\newblock Modelingtoolkit: A composable graph transformation system for
  equation-based modeling, 2021{\natexlab{c}}.

\bibitem[Mart{\'\i}n-Guerrero \& Lamata(2021)Mart{\'\i}n-Guerrero and
  Lamata]{martin2021reinforcement}
Mart{\'\i}n-Guerrero, J.~D. and Lamata, L.
\newblock Reinforcement learning and physics.
\newblock \emph{Applied Sciences}, 11\penalty0 (18):\penalty0 8589, 2021.

\bibitem[Metz et~al.(2021)Metz, Freeman, Schoenholz, and
  Kachman]{metz2021gradients}
Metz, L., Freeman, C.~D., Schoenholz, S.~S., and Kachman, T.
\newblock Gradients are not all you need.
\newblock \emph{arXiv preprint arXiv:2111.05803}, 2021.

\bibitem[Misra(2019)]{misra2019mish}
Misra, D.
\newblock Mish: A self regularized non-monotonic activation function.
\newblock \emph{arXiv preprint arXiv:1908.08681}, 2019.

\bibitem[Pearlmutter(1995)]{pearlmutter1995gradient}
Pearlmutter, B.~A.
\newblock Gradient calculations for dynamic recurrent neural networks: A
  survey.
\newblock \emph{IEEE Transactions on Neural networks}, 6\penalty0 (5):\penalty0
  1212--1228, 1995.

\bibitem[Polosecki et~al.(2020)Polosecki, Castro, Rish, Pustina, Warner, Wood,
  Sampaio, and Cecchi]{polosecki2020resting}
Polosecki, P., Castro, E., Rish, I., Pustina, D., Warner, J.~H., Wood, A.,
  Sampaio, C., and Cecchi, G.~A.
\newblock Resting-state connectivity stratifies premanifest huntington’s
  disease by longitudinal cognitive decline rate.
\newblock \emph{Scientific reports}, 10\penalty0 (1):\penalty0 1--15, 2020.

\bibitem[Rackauckas \& Nie(2017)Rackauckas and
  Nie]{rackauckas2017differentialequations}
Rackauckas, C. and Nie, Q.
\newblock Differentialequations.jl--a performant and feature-rich ecosystem for
  solving differential equations in julia.
\newblock \emph{Journal of Open Research Software}, 5\penalty0 (1):\penalty0
  15, 2017.

\bibitem[Rackauckas et~al.(2019)Rackauckas, Innes, Ma, Bettencourt, White, and
  Dixit]{rackauckas2019diffeqflux}
Rackauckas, C., Innes, M., Ma, Y., Bettencourt, J., White, L., and Dixit, V.
\newblock Diffeqflux.jl-a julia library for neural differential equations.
\newblock \emph{arXiv preprint arXiv:1902.02376}, 2019.

\bibitem[Rackauckas et~al.(2020)Rackauckas, Ma, Martensen, Warner, Zubov,
  Supekar, Skinner, Ramadhan, and Edelman]{rackauckas_2020}
Rackauckas, C., Ma, Y., Martensen, J., Warner, C., Zubov, K., Supekar, R.,
  Skinner, D., Ramadhan, A., and Edelman, A.
\newblock Universal differential equations for scientific machine learning,
  2020.
\newblock URL \url{https://arxiv.org/abs/2001.04385}.

\bibitem[Raissi et~al.(2019)Raissi, Perdikaris, and Karniadakis]{RAISSI2019686}
Raissi, M., Perdikaris, P., and Karniadakis, G.
\newblock Physics-informed neural networks: A deep learning framework for
  solving forward and inverse problems involving nonlinear partial differential
  equations.
\newblock \emph{Journal of Computational Physics}, 378:\penalty0 686--707,
  2019.
\newblock ISSN 0021-9991.
\newblock \doi{https://doi.org/10.1016/j.jcp.2018.10.045}.
\newblock URL
  \url{https://www.sciencedirect.com/science/article/pii/S0021999118307125}.

\bibitem[Ramezanian-Panahi et~al.(2022)Ramezanian-Panahi, Abrevaya,
  Gagnon-Audet, Voleti, Rish, and Dumas]{ramezanian2022generative}
Ramezanian-Panahi, M., Abrevaya, G., Gagnon-Audet, J.-C., Voleti, V., Rish, I.,
  and Dumas, G.
\newblock Generative models of brain dynamics.
\newblock \emph{Frontiers in Artificial Intelligence}, pp.\  147, 2022.

\bibitem[Ribeiro et~al.(2020)Ribeiro, Tiels, Umenberger, Sch{\"o}n, and
  Aguirre]{ribeiro2020smoothness}
Ribeiro, A.~H., Tiels, K., Umenberger, J., Sch{\"o}n, T.~B., and Aguirre, L.~A.
\newblock On the smoothness of nonlinear system identification.
\newblock \emph{Automatica}, 121:\penalty0 109158, 2020.

\bibitem[Rico-Martinez \& Kevrekidis(1993)Rico-Martinez and
  Kevrekidis]{rico1993continuous}
Rico-Martinez, R. and Kevrekidis, I.~G.
\newblock Continuous time modeling of nonlinear systems: A neural network-based
  approach.
\newblock In \emph{IEEE International Conference on Neural Networks}, pp.\
  1522--1525. IEEE, 1993.

\bibitem[Rico-Martinez et~al.(1992)Rico-Martinez, Krischer, Kevrekidis, Kube,
  and Hudson]{rico1992discrete}
Rico-Martinez, R., Krischer, K., Kevrekidis, I., Kube, M., and Hudson, J.
\newblock Discrete-vs. continuous-time nonlinear signal processing of cu
  electrodissolution data.
\newblock \emph{Chemical Engineering Communications}, 118\penalty0
  (1):\penalty0 25--48, 1992.

\bibitem[Rico-Martinez et~al.(1994)Rico-Martinez, Anderson, and
  Kevrekidis]{rico1994continuous}
Rico-Martinez, R., Anderson, J., and Kevrekidis, I.
\newblock Continuous-time nonlinear signal processing: a neural network based
  approach for gray box identification.
\newblock In \emph{Proceedings of IEEE Workshop on Neural Networks for Signal
  Processing}, pp.\  596--605. IEEE, 1994.

\bibitem[Rubanova et~al.(2019)Rubanova, Chen, and Duvenaud]{rubanova2019latent}
Rubanova, Y., Chen, R.~T., and Duvenaud, D.~K.
\newblock Latent ordinary differential equations for irregularly-sampled time
  series.
\newblock \emph{Advances in neural information processing systems}, 32, 2019.

\bibitem[Shen et~al.(2023)Shen, Appling, Gentine, Bandai, Gupta, Tartakovsky,
  Baity-Jesi, Fenicia, Kifer, Li, et~al.]{shen2023differentiable}
Shen, C., Appling, A.~P., Gentine, P., Bandai, T., Gupta, H., Tartakovsky, A.,
  Baity-Jesi, M., Fenicia, F., Kifer, D., Li, L., et~al.
\newblock Differentiable modelling to unify machine learning and physical
  models for geosciences.
\newblock \emph{Nature Reviews Earth \& Environment}, pp.\  1--16, 2023.

\bibitem[Turan \& J{\"a}schke(2021)Turan and J{\"a}schke]{turan2021multiple}
Turan, E.~M. and J{\"a}schke, J.
\newblock Multiple shooting for training neural differential equations on time
  series.
\newblock \emph{IEEE Control Systems Letters}, 6:\penalty0 1897--1902, 2021.

\bibitem[Varoquaux et~al.(2010)Varoquaux, Sadaghiani, Pinel, Kleinschmidt,
  Poline, and Thirion]{varoquaux2010group}
Varoquaux, G., Sadaghiani, S., Pinel, P., Kleinschmidt, A., Poline, J.-B., and
  Thirion, B.
\newblock A group model for stable multi-subject ica on fmri datasets.
\newblock \emph{Neuroimage}, 51\penalty0 (1):\penalty0 288--299, 2010.

\bibitem[Vaswani et~al.(2017)Vaswani, Shazeer, Parmar, Uszkoreit, Jones, Gomez,
  Kaiser, and Polosukhin]{vaswani2017attention}
Vaswani, A., Shazeer, N., Parmar, N., Uszkoreit, J., Jones, L., Gomez, A.~N.,
  Kaiser, {\L}., and Polosukhin, I.
\newblock Attention is all you need.
\newblock \emph{Advances in neural information processing systems}, 30, 2017.

\bibitem[Viquerat et~al.(2022)Viquerat, Meliga, Larcher, and
  Hachem]{viquerat2022review}
Viquerat, J., Meliga, P., Larcher, A., and Hachem, E.
\newblock A review on deep reinforcement learning for fluid mechanics: An
  update.
\newblock \emph{Physics of Fluids}, 34\penalty0 (11), 2022.

\bibitem[von Rueden et~al.(2023)von Rueden, Mayer, Beckh, Georgiev,
  Giesselbach, Heese, Kirsch, Pfrommer, Pick, Ramamurthy, Walczak, Garcke,
  Bauckhage, and Schuecker]{von_rueden_2021}
von Rueden, L., Mayer, S., Beckh, K., Georgiev, B., Giesselbach, S., Heese, R.,
  Kirsch, B., Pfrommer, J., Pick, A., Ramamurthy, R., Walczak, M., Garcke, J.,
  Bauckhage, C., and Schuecker, J.
\newblock Informed machine learning – a taxonomy and survey of integrating
  prior knowledge into learning systems.
\newblock \emph{IEEE Transactions on Knowledge and Data Engineering},
  35\penalty0 (1):\penalty0 614--633, 2023.
\newblock \doi{10.1109/TKDE.2021.3079836}.

\bibitem[Wang et~al.(2021)Wang, Teng, and Perdikaris]{wang_siam_2021}
Wang, S., Teng, Y., and Perdikaris, P.
\newblock Understanding and mitigating gradient flow pathologies in
  physics-informed neural networks.
\newblock \emph{SIAM Journal on Scientific Computing}, 43\penalty0
  (5):\penalty0 A3055--A3081, 2021.
\newblock \doi{10.1137/20M1318043}.
\newblock URL \url{https://doi.org/10.1137/20M1318043}.

\bibitem[Weinan(2017)]{weinan2017proposal}
Weinan, E.
\newblock A proposal on machine learning via dynamical systems.
\newblock \emph{Communications in Mathematics and Statistics}, 1\penalty0
  (5):\penalty0 1--11, 2017.

\bibitem[Willard et~al.(2022)Willard, Jia, Xu, Steinbach, and
  Kumar]{willard_2022}
Willard, J., Jia, X., Xu, S., Steinbach, M., and Kumar, V.
\newblock Integrating scientific knowledge with machine learning for
  engineering and environmental systems.
\newblock \emph{ACM Comput. Surv.}, 55\penalty0 (4), nov 2022.
\newblock ISSN 0360-0300.
\newblock \doi{10.1145/3514228}.
\newblock URL \url{https://doi.org/10.1145/3514228}.

\end{thebibliography}
\bibliographystyle{icml2021}

\newpage

\section*{Supplementary Information}
\renewcommand{\thesubsection}{\Alph{subsection}}
\renewcommand \thesubsubsection {\thesubsection.\arabic{subsubsection}}

\subsection{Models architectures}
Referring to the diagram in Figure \ref{fig:LatentDiffeq_schema}, the specific architecture used for the different models, for both simulated and empirical data experiments, is as follows:

\subsubsection{Basic GOKU-nets}

\paragraph{Feature Extractor}\quad

ResNet with 4 fully-connected layers, each with 200 neurons and using mish activation functions~\citep{misra2019mish}. Input dimension = number of dimensions in the input data. Output dimension = 128.

\paragraph{Pattern Extractor}\quad

Initial values path: an RNN with 2 layers and 64 neurons in each with ReLU activations. Input dimension = 128. Output dimension = 64.

Parameters path: Bidirectional LSTM with 2 layers and 64 neurons in each. Input dimension = 128. Output dimension = 128. Note that the dimension of the output of the forward LSTM and the backward LSTM are 64 but when concatenating them, the resulting output dimension is the given one.

\paragraph{Latent in}\quad

Initial values path: single-layered fully connected NN. Input dimension = 64. Output dimension = 64.

Parameters path: fully connected NN with 1 layer. Input dimension = 128. Output dimension = 128.

\paragraph{Latent out}\quad

Initial values path: fully connected NN with 2 layers and 200 neurons in the hidden layer, using no activation function (identity). Input dimension = 64. Output dimension = number of state variables of the differential equation.

Parameters path: fully connected NN with 2 layers and 200 neurons in the hidden layer, using sigmoid activation function. The parameters are projected from the interval [0, 1] to the desired range when integrating the differential equation. Input dimension = 128. Output dimension = number of parameters of the differential equation.

\paragraph{Differential Equation layer}\quad

The predefined differential equation is solved numerically for each of the sets of parameters and initial conditions provided by the previous layer. The output is the trajectories at time points equivalent to the input data.

\paragraph{Reconstructor}\quad

ResNet is similar to the one in the Feature Extractor, except that in this case the input dimension is the number of state variables of the differential equation and the output dimension is the one corresponding to the input data.

\subsubsection{GOKU-nets with attention}

With the exception of the Pattern Extractor, the rest of the layers in the GOKU-nets with attention model remain identical to those in the basic GOKU-nets.

\paragraph{Pattern Extractor}\quad

Initial values path: LSTM with 1 layer. Input dimension = 128. Output dimension = 128.

Parameters path: Bidirectional LSTM (BiLSTM) with 1 layer. Input dimension = 128. Output dimension = 128. A fully connected NN with input and output dimensions of 128 is used for the attention mechanism. This attention NN processes all the output sequences of the BiLSTM, after which a softmax is applied across the time dimension in order to obtain the attentional scores that will be used in the weighted sum of all the time steps returned by the BiLSTM.

\subsubsection{LSTM baseline model}

The whole architecture is the same as in the basic GOKU-net, except for the Differential Equation layer, which is replaced by an LSTM:

\paragraph{LSTM layer}\quad

We used a single-layered LSTM with input and output dimensions set to $z\_dim$. This value is determined in each experiment to ensure that the total number of parameters in the LSTM model closely matches that of the corresponding GOKU-UI. For the simulated dataset experiments, we set $z\_dim = 42$. In the case of the empirical dataset experiments, $z\_dim = 105$. The LSTM operates recursively. It takes as its first input the value equivalent to the initial condition in differential equations. Subsequently, the model feeds back its last output as the new input, continuing this process until the number of time steps matches that of the model's input.

\subsubsection{Latent ODE baseline model}

The whole architecture is the same as in the basic GOKU-net, except for the Differential Equation layer, which is replaced by a Neural ODE:

\paragraph{Neural ODE layer}\quad

Neural ODE is parametrized by a fully connected NN with 3 layers and $node\_hidden\_dim$ neurons in each. The input and output dimensions are given by $z\_dim$, which is the number of state variables. In the case of the simulated dataset experiments, the number of state variables was selected to match the true latent dimension $z\_dim = 6$ and the number of neurons in each layer was adjusted so that the total number of parameters in the model matched as closely as possible that of the corresponding GOKU-UI, resulting in $node\_hidden\_dim = 137$. On the other hand, in the case of the fMRI experiments, the number of state variables was set to $z\_dim = 20$ and $node\_hidden\_dim = 317$, also matching the total number of parameters of the corresponding GOKU-UI model.

\subsection{Comprehensive description of experiments}
\subsubsection{Simulated dataset generation}
The high-dimensional simulated dataset used for training the model was constructed based on the simulations of 3 coupled Stuart-Landau oscillators (Eqs. \ref{eq:stuart-landau_network}) with different random sets of parameters. Each set of parameters corresponds to a different training sample. Whenever we used the Stuart-Landau model in our experiments (both when generating the dataset and when using it inside the GOKU-nets), the time was rescaled by multiplying the right-hand side of Eqs. \ref{eq:stuart-landau_network} by 20. Thus, when integrating the equations with the used $dt=0.05$, the input sequences of length 46 time steps contain a few oscillations. The parameters $a$, $\omega$ and $C$ were sampled from uniform distributions within the following ranges

$a \in [-0.2, 0.2]$

$\omega \in [0.08 \pi, 0.14 \pi]$

$C \in [0, 0.2]$

while $G = 0.1$ and $\eta = 0.02$. On the other hand, the initial conditions for the six state variables were sampled from uniform distributions within the ranges $[0.3, 0.4]$. For each set of parameters and initial conditions, the system is integrated with the {\fontfamily{cmtt}\selectfont SOSRI} solver, a Stability-optimized adaptive strong order 1.5 and weak order 2.0 for diagonal/scalar Ito SDEs, from the DifferentialEquations.jl Julia package~\citep{rackauckas2017differentialequations}. The complete time span of the integration is 35 units of time and the trajectories are saved every 0.05, resulting in 700 time points. The first 100 time steps are trimmed, in order to remove possible initial transients. Afterwards, a random linear transformation is independently applied to each of the 600 remaining time steps, in order to obtain 784 dimensions. In other words, every state vector of length 6 from each sample is multiplied by the same 784×6 matrix, initialized randomly sampling from a uniform distribution in the range [-1, 1]. A training dataset was created with 5000 samples, which serves as the source for the different training instances using different sizes of training sets (see Figure \ref{fig:Hopf_data_scaling}). A different test set with 900 samples was created for the posterior evaluations of the model.

\subsubsection{Empirical dataset generation}

We used resting state fMRI data from 153 participants, obtained from the Track-On HD study~\citep{kloppel2015compensation}. The data underwent pre-processing, as described in~\citet{polosecki2020resting}, and a 20-component Canonical ICA~\citep{varoquaux2010group} was performed. Upon inspecting the resulting 20 components, 9 were identified as artifacts and thus discarded, leaving 11 components for further analysis in our experiments. Each subject contributed data from two visits, accumulating a total of 306 data samples. Each sample comprised 160 time points, obtained at a temporal resolution of 3 seconds.

For our investigation, we set aside approximately 20\% of the data samples (n=60) for testing, while ensuring balanced representation from sex, condition, and measurement site. The remaining data samples (n=246) were allocated for training and validation. Specifically, the first 114 time points from each of these samples were utilized for model training, with the remainder reserved for validation and early training termination. Finally, the training, validation, and test splits were all normalized by the standard deviation of the training set.

\subsubsection{Training settings}

All the experiments underwent the same training procedure with identical hyperparameters, which will be described here.

The input sequence length for all the models was 46 time steps, and the batch size was set at 64. As described above, the full length of each sample in the training sets was 600 time steps for the synthetic dataset and 114 for the fMRI dataset. The procedure for generating a batch of training data is as follows: First, 64 samples that have not been used previously in the current training epoch are randomly selected. Then, for each sample, a 46 time-step-long interval is randomly chosen within the 600 or 114 time steps available in the full sample length.

The GOKU-net based models, contain the same Stuart-Landau differential equations as described above, however, the allowed ranges of parameters differ from the ones used during the generation of the synthetic dataset. In order to be closer to a real world use-case we allow for a wider range of parameters than those actually used for generating the data, since in principle one would not know the true range:

$a \in [-1, 1]$

$\omega \in [0, 1]$

while keeping, the other parameters the same except of the connectivity in the empirical fMRI training, in which case it was allowed to be negative: $C \in [-0.2, 0.2]$. The differential equations definitions were optimized for higher computational performance with the help of ModelingToolkit.jl~\citep{ma2021modelingtoolkit}. During training, they were solved with the {\fontfamily{cmtt}\selectfont SOSRI} solver, a Stability-optimized adaptive strong order 1.5 and weak order 2.0 for diagonal/scalar Ito SDEs, from the DifferentialEquations.jl Julia package~\citep{rackauckas2017differentialequations}. The sensitivity algorithm used was {\fontfamily{cmtt}\selectfont ForwardDiffSensitivity} from the SciMLSensitivity.jl package~\citep{rackauckas_2020}.
The models were defined and trained within the deep learning framework of the Flux.jl package~\citep{Flux.jl-2018}. The experiments were managed using DrWatson.jl package~\citep{Datseris2020}.

\begin{figure}[!ht]
    \centering
    \includegraphics[width=0.8\linewidth]{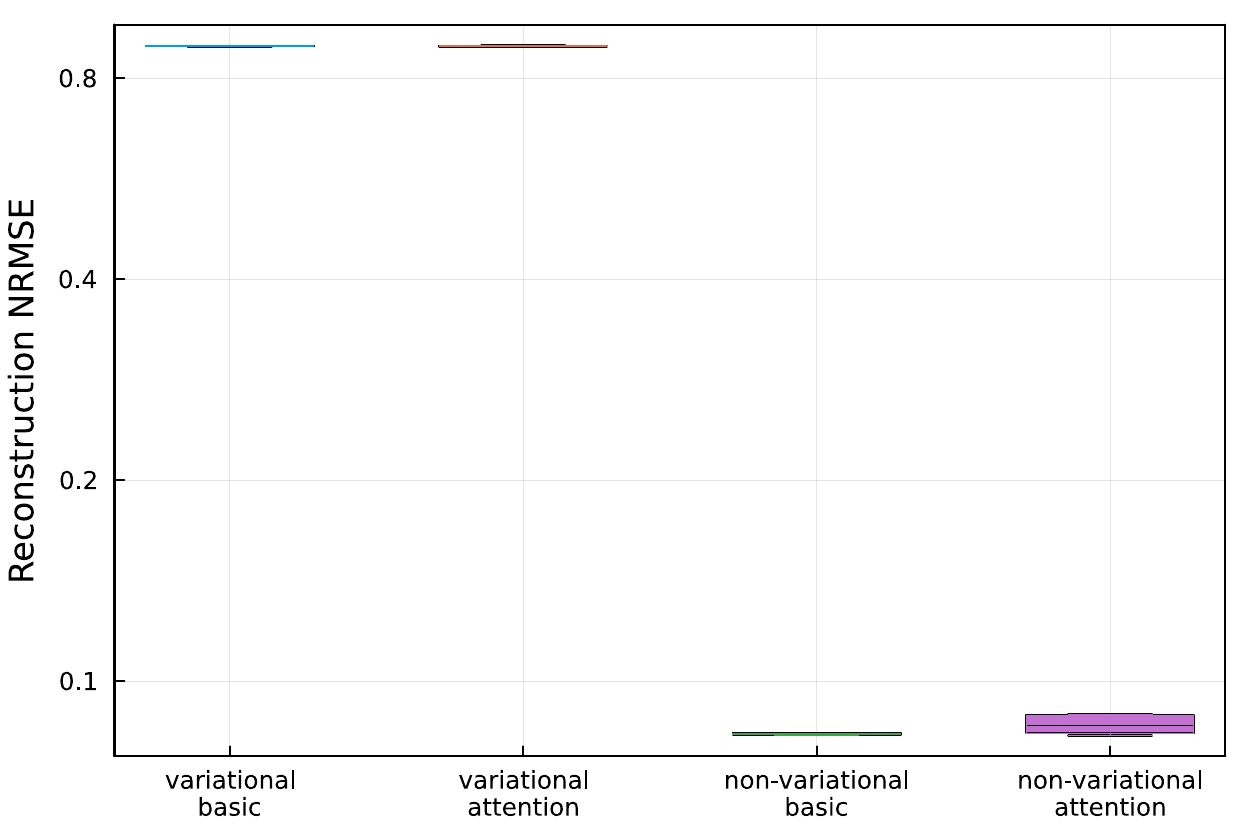}
    \caption{Box plot of the normalized RMSE on the test synthetic dataset for GOKU-nets trained with multiple shooting either in their basic or attention variants and in a variational or non-variational version. All hyperparameters are kept as in the main paper experiments.
    }
    \label{fig:variational}
\end{figure}

The model was trained with Adam with a weight decay of $10^{-10}$, and the learning rate was dynamically determined by the following schedule. The learning rate begins with a linear growth (also referred to as \emph{learning rate warm-up}) from $10^{-7}$, escalating up to 0.005251 across 20 epochs. Afterwards, it maintains that value until the validation loss stagnates (has not achieved a lower value for 50 epochs), at which point it starts a sinusoidal schedule with an exponentially decreasing amplitude. 

For the multiple shooting training, all the presented experiments used a time window length of 10, therefore partitioning 46-time-steps-long sequences into 5 windows with their endpoints overlapping. The regularization coefficient in the loss function for the continuity constraint had a value of 2.

Since we found that models with variational autoencoders underperform their non-variational versions (see Figure \ref{fig:variational}), all the results presented in this work were obtained using non-variational GOKU-nets. This is, instead of sampling from normal distributions in the latent space as depicted in Figure \ref{fig:LatentDiffeq_schema}, we pass forward the mean values $\mu_{z_0}$ and $\mu_{\theta}$. Thus, the associated loss function does not have the KL divergence term associated with the ELBO but retains the reconstruction loss given by the mean squared error between the output of the model and the input, normalized by the mean absolute value of the input. In addition, when multiple shooting training is employed, the extra term regarding the continuity constraint is included in the loss function. This extra term consists of the mean squared differences between the last point of a window and the initial from the next one, divided by the number of junctions and multiplied by a regularization coefficient. Please, note that this continuity regularization is performed in the state space of the differential equation and not in the input space.

\subsection{Further exploration of the model}
\label{sec:supplemental-further}

In Figure \ref{fig:variational}, a box plot is presented, comparing the reconstruction performances on the synthetic dataset when the models are variational versus when they are not.

\begin{figure}[ht]
    \centering
    \includegraphics[width=0.8\linewidth]{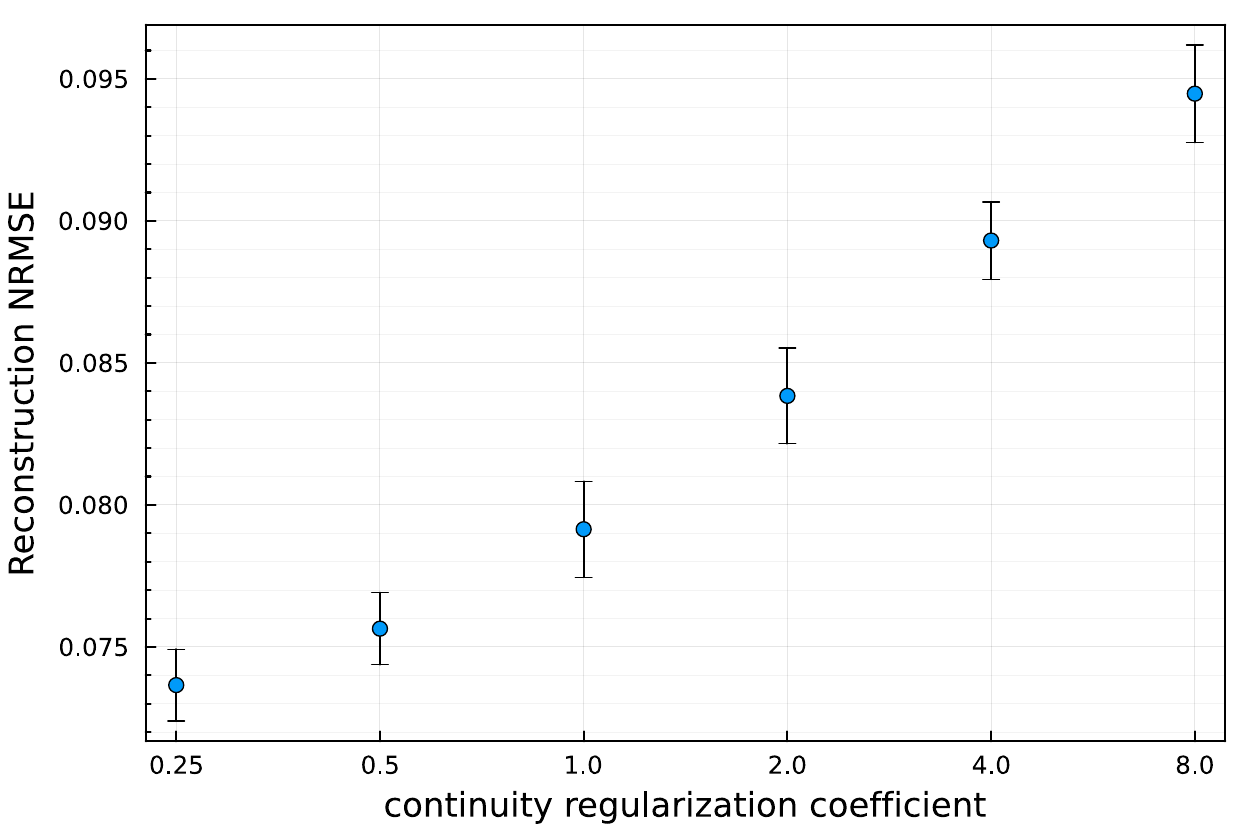}
    \caption{ Reconstruction normalized RMSE of GOKU-UI on the test synthetic dataset for different values of the regularization coefficient in the loss function regarding the continuity constraint. All other hyperparameters are kept as in the main paper experiments.
    }
    \label{fig:cont_term_sweep}
\end{figure}

\begin{figure}[ht]
    \centering
    \includegraphics[width=0.8\linewidth]{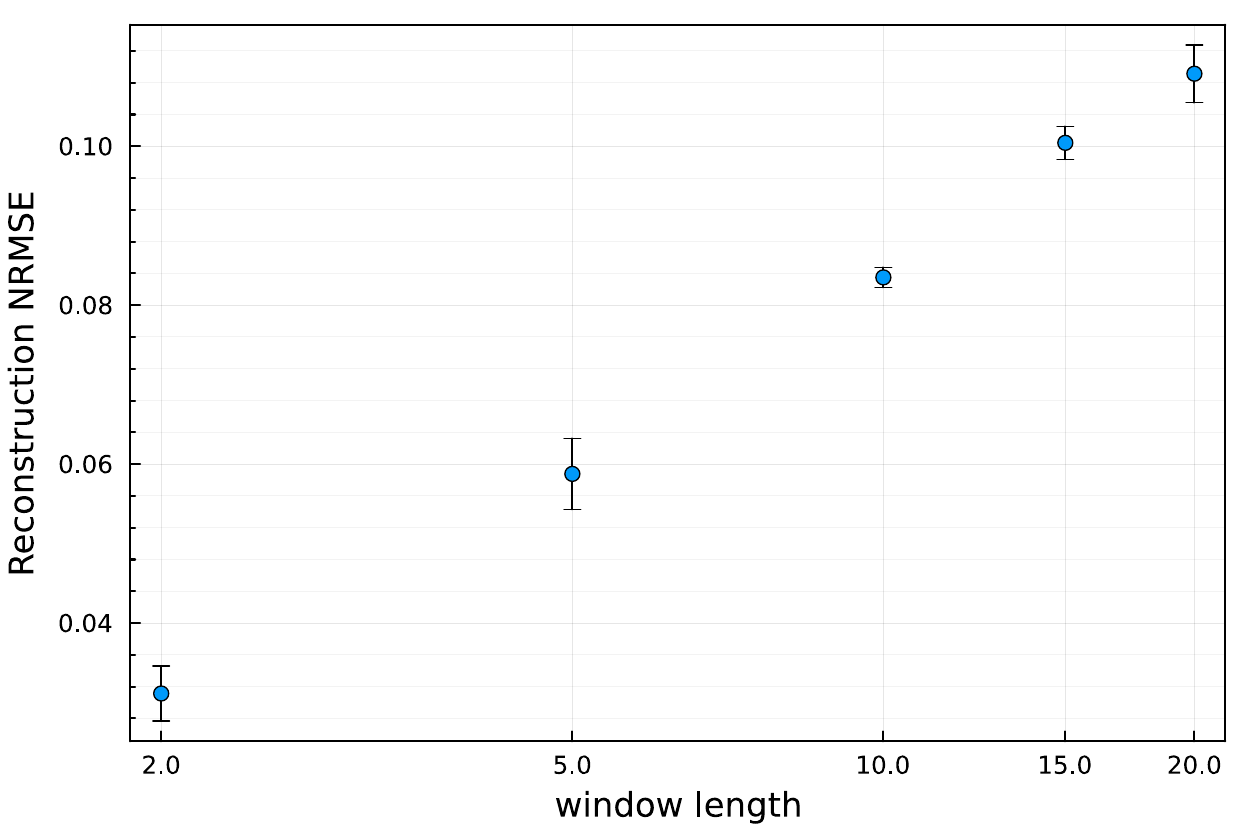}
    \caption{ Reconstruction normalized RMSE of GOKU-UI on the test synthetic dataset for different values of time window length for the multiple shooting partition. All other hyperparameters are kept as in the main paper experiments.
    }
    \label{fig:win_len_sweep}
\end{figure}

Figures \ref{fig:cont_term_sweep} and \ref{fig:win_len_sweep} present the performance of the GOKU-UI on the synthetic dataset when trained using different hyperparameters. As a reminder, the base values used in our experiments for the continuity regularization coefficient, window length and input sequence length are 2, 10 and 46 respectively. In Figure \ref{fig:cont_term_sweep} all hyperparameters are kept the same, except for the continuity coefficient. Similarly, Figure \ref{fig:win_len_sweep} shows variations with respect to the window length and Figure \ref{fig:win_len_sweep}, respect to the input sequence length.

\begin{figure}[h]
    \centering
    \includegraphics[width=0.8\linewidth]{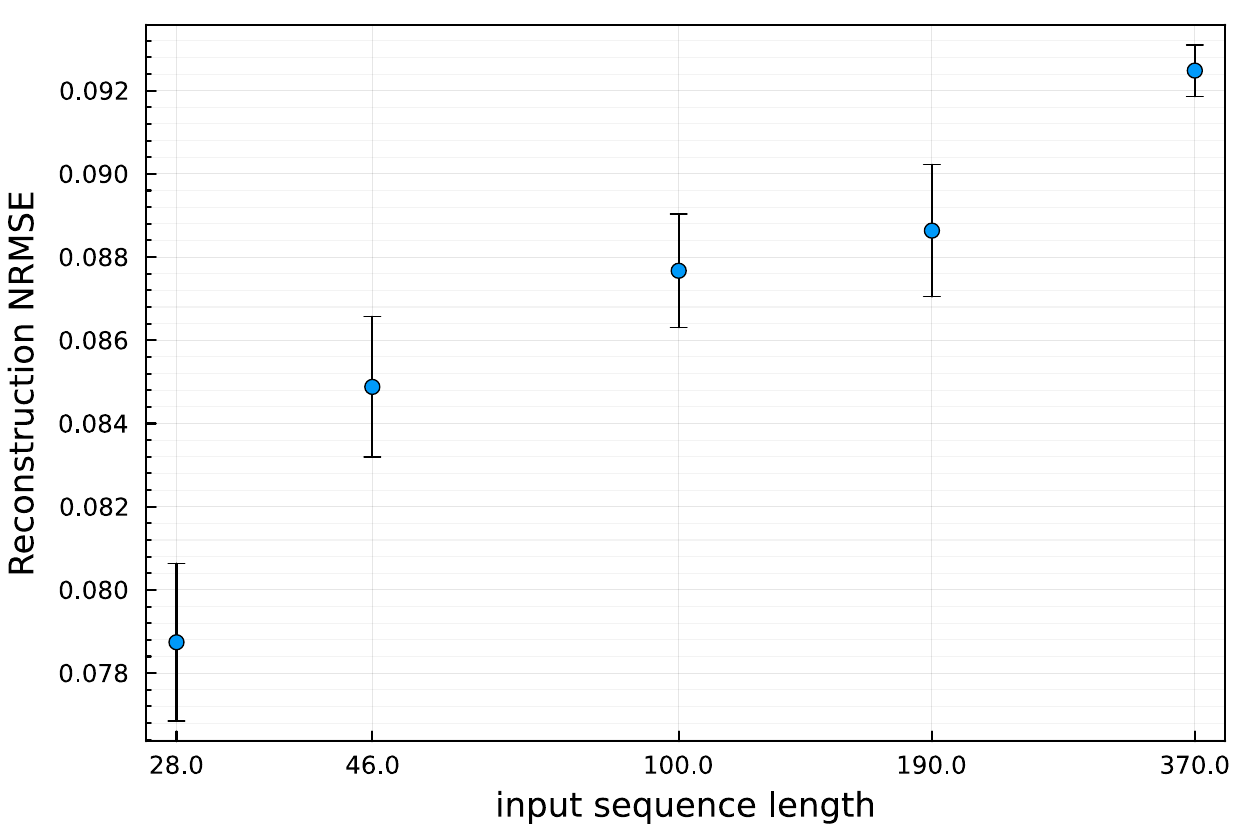}
    \caption{ Reconstruction normalized RMSE of GOKU-UI on the test synthetic dataset for different values of time input sequence length. All other hyperparameters are kept as in the main paper experiments.
    }
    \label{fig:seq_len_sweep}
\end{figure}

As evidenced by these results, there exist other sets of hyperparameters that produce higher performances than those obtained with the base hyperparameters used in our experiments. The set of hyperparameters had been selected based on a grid search but while using a different dynamical system. In any case, the results presented in this paper serve as a proof of concept, showing that even when using a sub-optimal set of hyperparameters, GOKU-UI demonstrates significantly better reconstruction and forecast performances with respect to the baselines models and other GOKU-nets variants tested.

\begin{figure}[ht]
    \centering
    \includegraphics[trim={0cm 2.15cm 0cm 0cm},clip, width=0.6\linewidth]{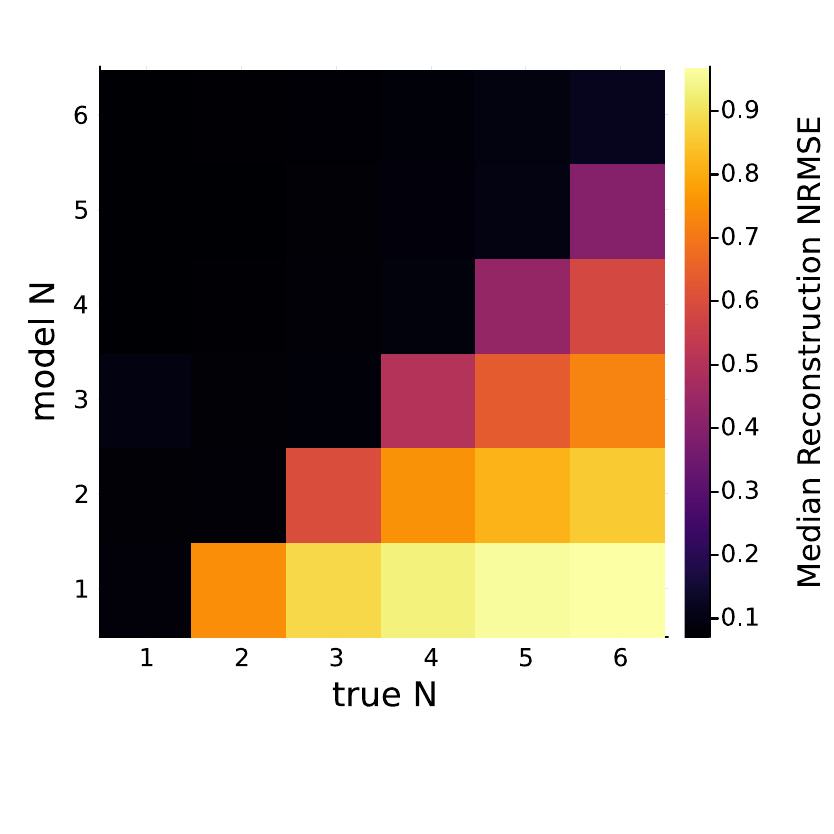}
    \caption{ Reconstruction errors of GOKU-UI using different number of oscillators (\emph{model N}) on synthetic Stuart-Landau datasets constructed with \emph{true N} number of oscillators.
    }
    \label{fig:hopf_data_dim_sweep}
\end{figure}

Motivated by the question of whether it would be possible to identify the latent dimensionality of some data using the GOKU-UI model, the following experiments were performed. GOKU-UI models with different numbers of Stuart-Landau oscillators were trained on synthetic datasets generated from distinct numbers of latent oscillators but all with the same input size of 784, so the latent dimensionality was not evident. Furthermore, as in the training settings from our other experiments, the allowed parameter ranges were wider in the GOKU-UI than when generating the datasets. The results of such experiments are presented in Figure \ref{fig:hopf_data_dim_sweep}, where color represents testing reconstruction errors for GOKU-UIs with \emph{model N} oscillators on datasets built with latent \emph{true N} oscillators. We see that with an increasing number of oscillators in the model, the error progressively diminishes until reaching the true number of latent oscillators in the data, and from there, the error gets abruptly reduced. Notably, when using more oscillators in the model than the true latent in the data, the model still learns equally well how to reconstruct it. However, the most salient feature is that in this simulated scenario, it is possible to identify the true latent dimensionality of the data as such of the number of oscillators in GOKU-UI for which the reconstruction error gets abruptly reduced.

\begin{figure}[h]
    \centering
    \includegraphics[width=1\linewidth]{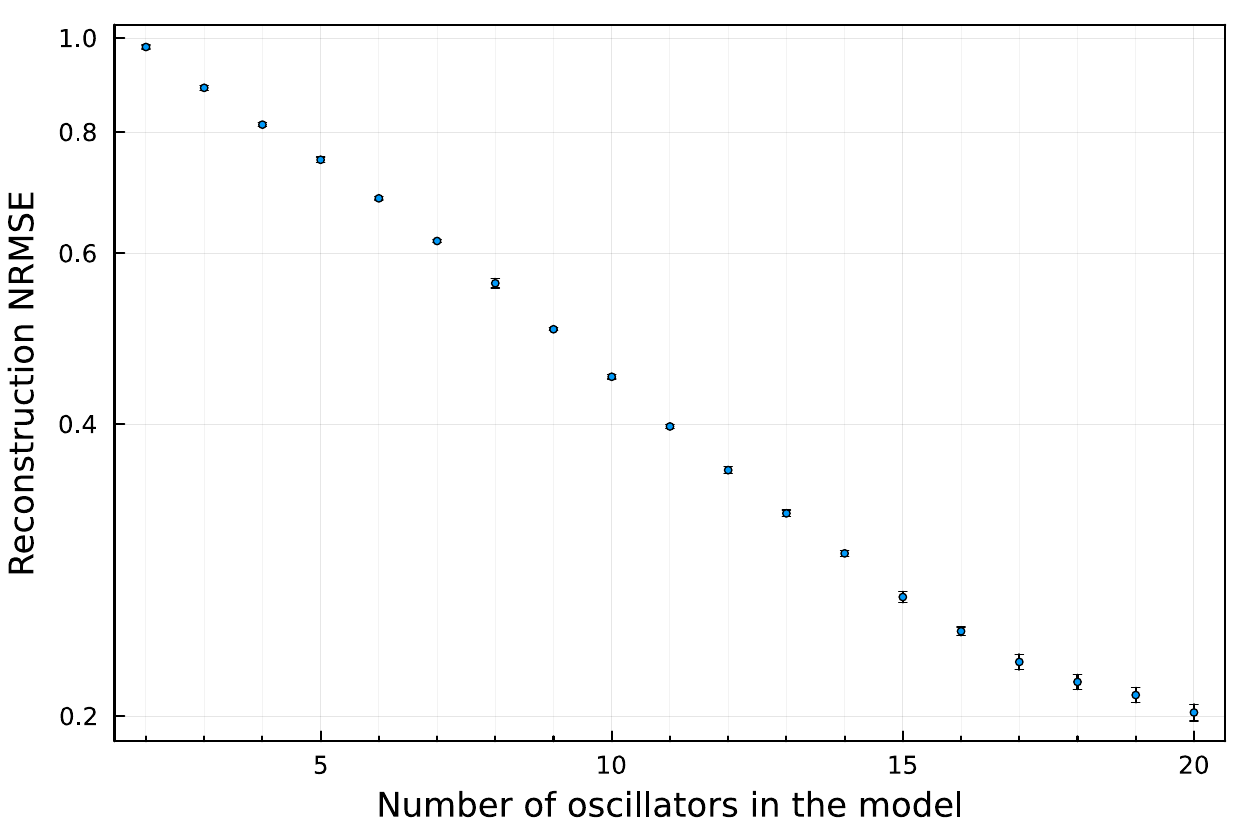}
    \caption{Reconstruction normalized RMSE on fMRI test data, while changing the number of Stuart Landau oscillators inside the GOKU-UI model.}
    \label{fig:fMRI_dim_sweep}
\end{figure}

A similar attempt was made to infer the latent dimensionality of the fMRI data. However, as seen in Figure~\ref{fig:fMRI_dim_sweep}(where the y-axis is in logarithmic scale), there is no abrupt descent in the reconstruction error for any number of oscillators inside GOKU-UI that we tried. Nevertheless, an inflection point can be identified for $N = 17$. We concluded that any $N \ge 17$ was adequate to use and chose $N = 20$ to perform the experiments presented in this study.

\subsection{Reconstruction plots}
\label{sec:rec_plots}
\begin{figure}[ht!]
    \centering
    \includegraphics[width=1\linewidth]{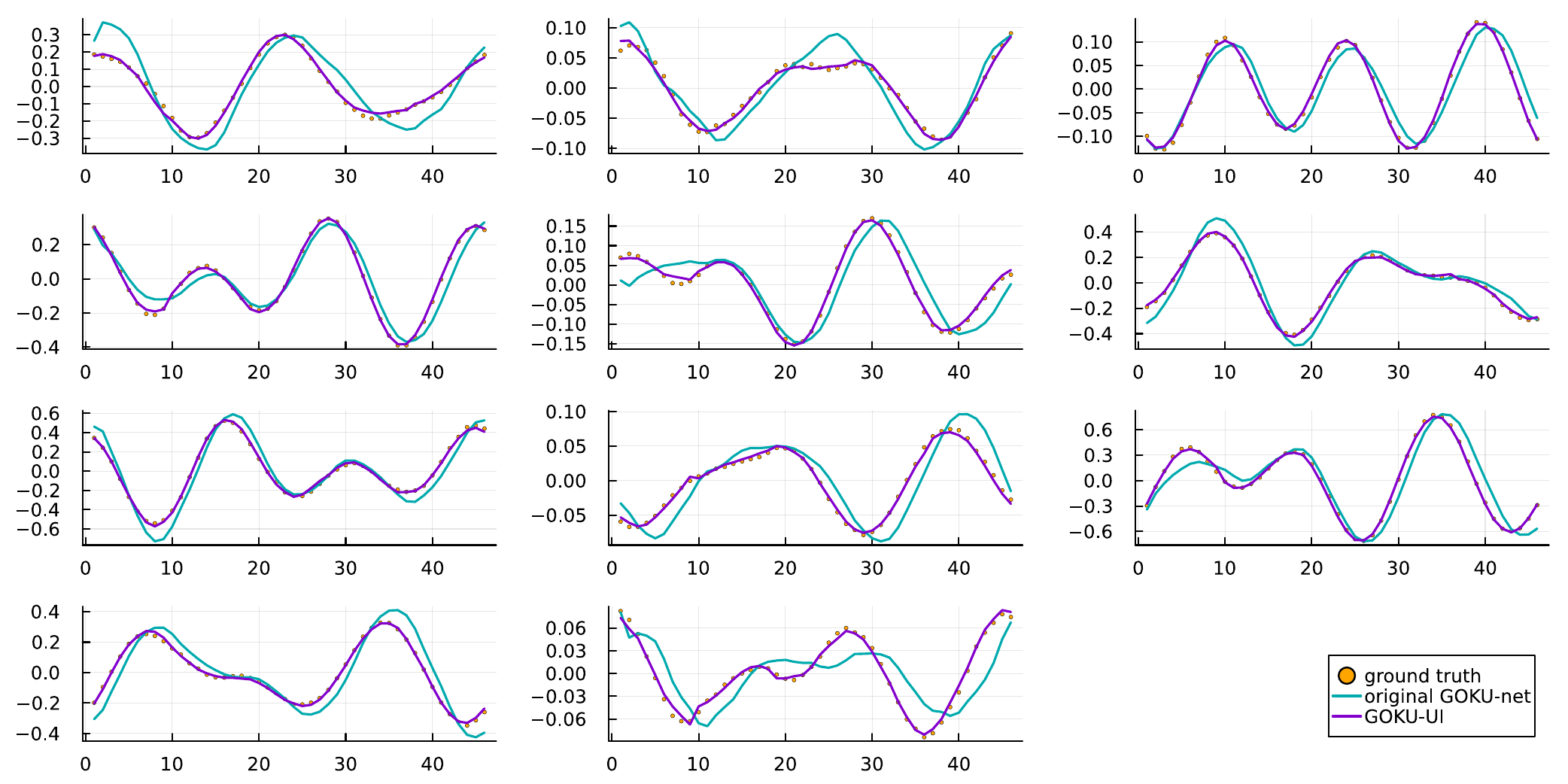}
    \caption{Representative example of a 46-time-step input sequence from the synthetic test set, accompanied by its reconstructions from both GOKU-UI and the original GOKU-nets (lacking attention mechanisms and trained with single shooting). The sample was selected so that its RMSE was the closest to the median error across all samples. 11 randomly selected components out of the 784 are displayed.
    }
    \label{fig:hopf_reconstruction_sample1}
\end{figure}

\begin{figure}[hb!]
    \centering
    \includegraphics[width=1\linewidth]{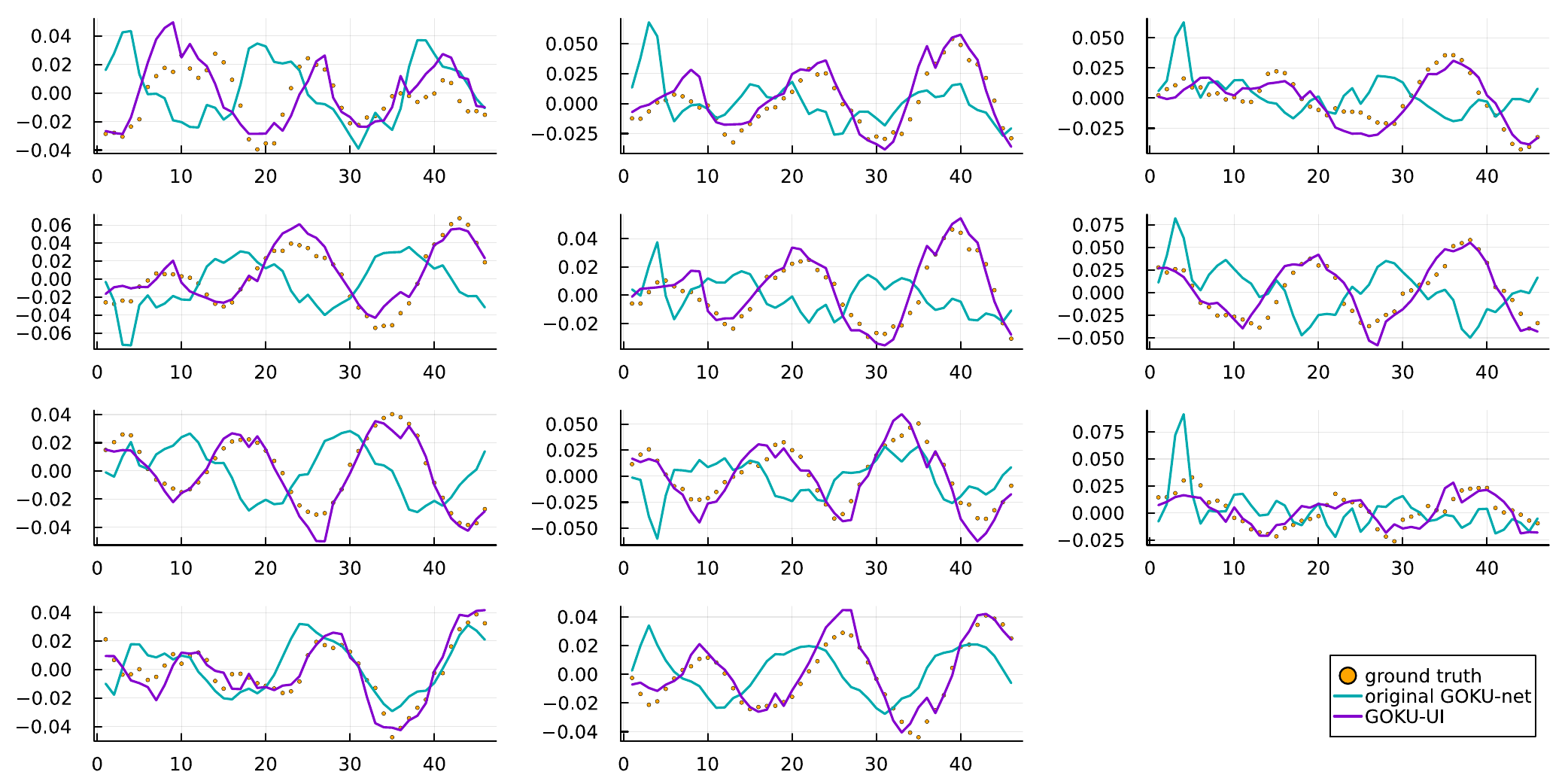}
    \caption{Representative example of a 46-time-step input sequence from the synthetic test set, accompanied by its reconstructions from both GOKU-UI and the original GOKU-nets (lacking attention mechanisms and trained with single shooting). The sample was selected so that its RMSE was the closest to the median error across all samples. 11 randomly selected components out of the 784 are displayed. This figure is similar to the previous one but presents results from different instances of the trained models, each initialized with a unique random seed.
    }
    \label{fig:hopf_reconstruction_sample2}
\end{figure}

\begin{figure}[ht!]
    \centering
    \includegraphics[width=1\linewidth]{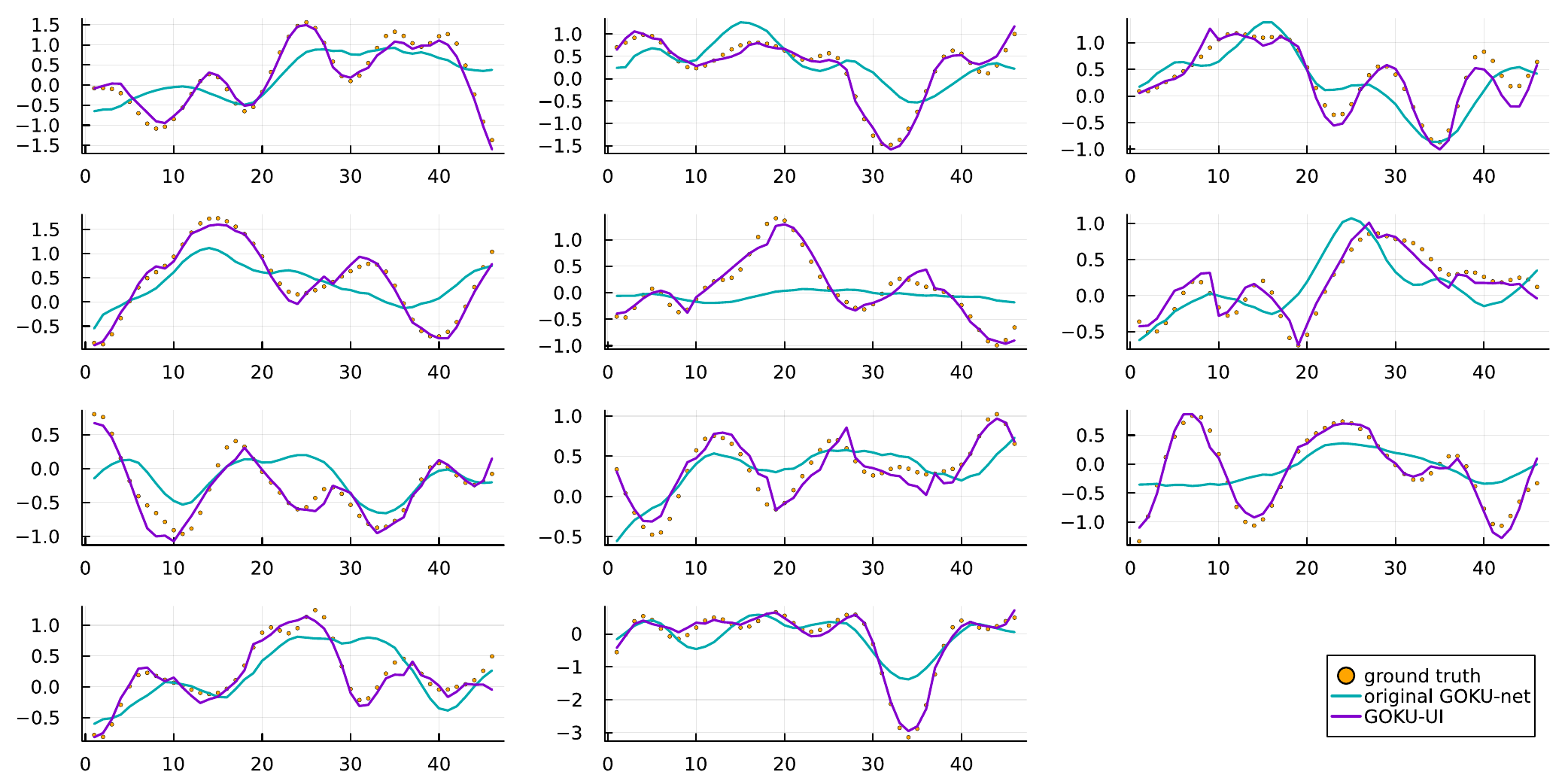}
    \caption{Representative example of a 46-time-steps input sequence for all considered ICA components from the empirical fMRI test set, accompanied by its reconstructions from both GOKU-UI and the original GOKU-nets (lacking attention mechanisms and trained with single shooting). The sample was selected so that its RMSE was closest to the median error across all samples. The x-axis represents time steps, each corresponding to 3 seconds.
    }
    \label{fig:fMRI_reconstruction_sample1}
\end{figure}

\begin{figure}[hb!]
    \centering
    \includegraphics[width=1\linewidth]{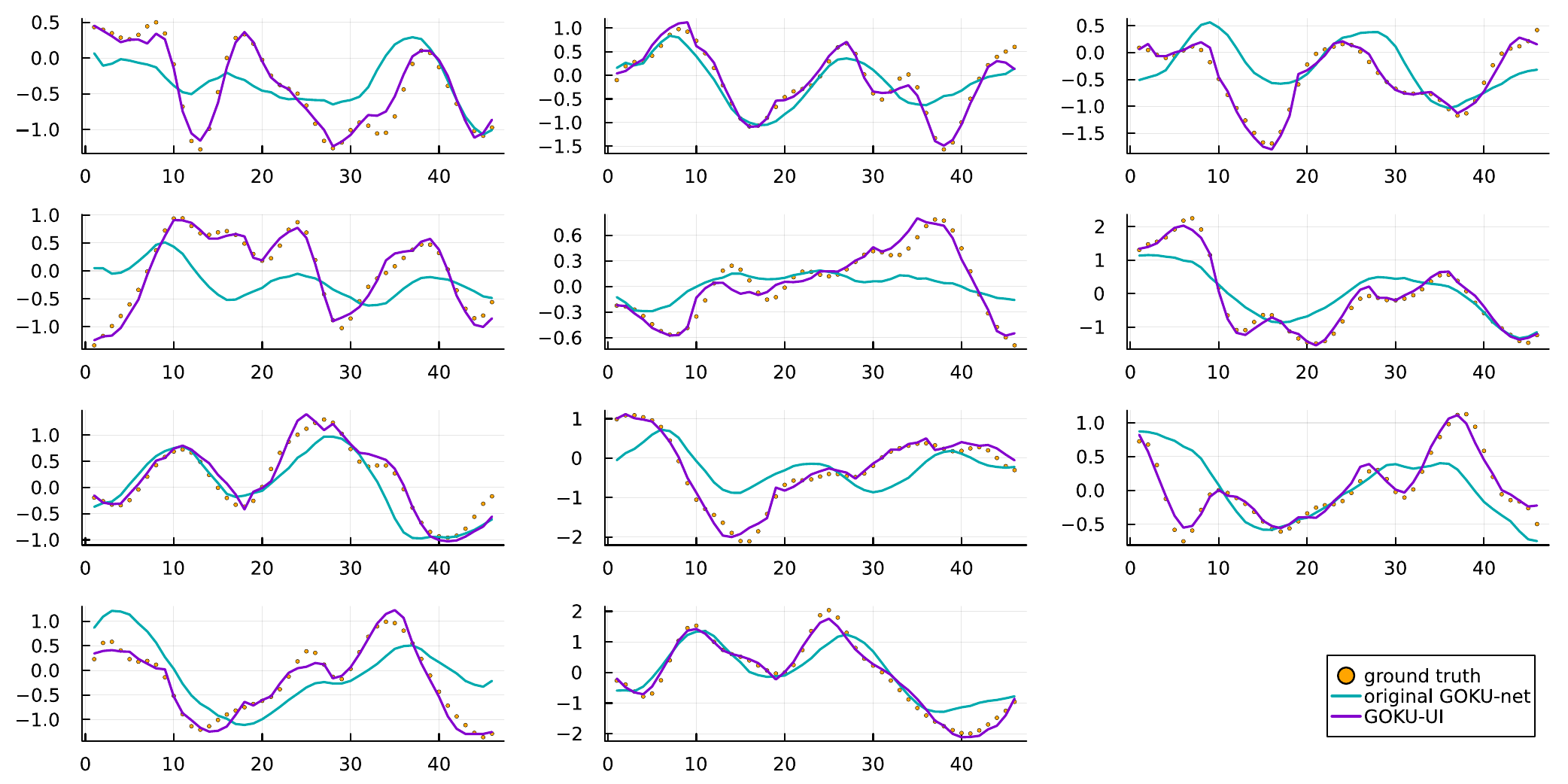}
    \caption{Representative example of a 46-time-steps input sequence for all considered ICA components from the empirical fMRI test set, accompanied by its reconstructions from both GOKU-UI and the original GOKU-nets (lacking attention mechanisms and trained with single shooting). The sample was selected so that its RMSE was closest to the median error across all samples. This figure is similar to the previous one but presents results from different instances of the trained models, each initialized with a unique random seed. The x-axis represents time steps, each corresponding to 3 seconds.
    }
    \label{fig:fMRI_reconstruction_sample2}
\end{figure}

To provide a visual representation of the model's performance, this section presents trajectories from both the synthetic and empirical fMRI test sets, along with their corresponding reconstructions by GOKU-UI and the original GOKU-nets (lacking attention mechanisms and trained with single shooting). The x-axis represents time steps in all cases. To display representative cases, samples were selected based on their mean reconstruction RMSE being closest to the median error across all samples. For the synthetic data, 11 components were randomly selected for display in Figures \ref{fig:hopf_reconstruction_sample1} and \ref{fig:hopf_reconstruction_sample2}, due to the impracticality of displaying all 784 components. Each figure displays results from different instances of models, all trained with 4800 samples but each initialized with a unique random seed. For the fMRI data, all 11 ICA components are displayed in Figures \ref{fig:fMRI_reconstruction_sample1} and \ref{fig:fMRI_reconstruction_sample2}.

\end{document}